\documentclass{article}

    \PassOptionsToPackage{numbers,sort&compress}{natbib}

\usepackage[preprint]{neurips_2024}

\usepackage[utf8]{inputenc} 
\usepackage[T1]{fontenc}    
\usepackage{hyperref}       
\usepackage{url}            
\usepackage{booktabs}       
\usepackage{amsfonts}       
\usepackage{nicefrac}       
\usepackage{microtype}      
\usepackage[table,xcdraw]{xcolor}

\usepackage{amsmath}
\DeclareMathOperator*{\argmax}{arg\,max}

\usepackage{bbm}
\usepackage{graphicx}

\usepackage{fixltx2e}
\usepackage{color}
\usepackage{algpseudocode}
\algtext*{EndWhile}
\algtext*{EndIf}
\algtext*{EndFunction}
\MakeRobust{\Call}
 
 \usepackage{algorithm}
 \usepackage{algorithmicx}

\algrenewcommand\algorithmicrequire{\textbf{Input:}}
\algrenewcommand\algorithmicensure{\textbf{Output:}}

\usepackage{multirow}
\usepackage{booktabs}
\usepackage{wrapfig}

\usepackage{caption} 
\usepackage{array,ragged2e}
\newcolumntype{R}[1]{>{\RaggedLeft\arraybackslash}p{#1}}

\usepackage{tabularx}
\usepackage{array} 

\usepackage{subcaption} 

\title{Ensemble Debiasing Across Class and Sample Levels for Fairer Prompting Accuracy}

\author{Ruixi Lin \qquad Ziqiao Wang \qquad Yang You\\
  Department of Computer Science\\
  National University of Singapore\\
  \texttt{
  \begin{tabular}{l}
ruixi@comp.nus.edu.sg \\ e1374492@u.nus.edu  \\ youy@comp.nus.edu.sg\\
\end{tabular}}\\}

\begin{document}

\maketitle

\begin{abstract}
Language models are strong few-shot learners and achieve good overall accuracy in text classification tasks, masking the fact that their results suffer from great class accuracy imbalance. We believe that the pursuit of overall accuracy should not come from enriching the strong classes, but from raising up the weak ones. To address the imbalance, we propose a Heaviside step function based ensemble debiasing method, which enables flexible rectifications of in-context learned class probabilities at both class and sample levels. Evaluations with Llama-2-13B on seven text classification benchmarks show that our approach achieves state-of-the-art overall accuracy gains with balanced class accuracies. More importantly, we perform analyses on the resulted probability correction scheme, showing that sample-level corrections are necessary to elevate weak classes. Due to effectively correcting weak classes, our method also brings significant performance gains to a larger model variant, Llama-2-70B, especially on a biomedical domain task, further demonstrating the necessity of ensemble debiasing at both levels. Our source code is available at \url{https://github.com/NUS-HPC-AI-Lab/DCS}.
\end{abstract}

\section{Introduction}
\label{sec:intro}
\begin{wrapfigure}[14]{R}{0.45\textwidth}
  \begin{center}
    \includegraphics[width=0.45\textwidth]{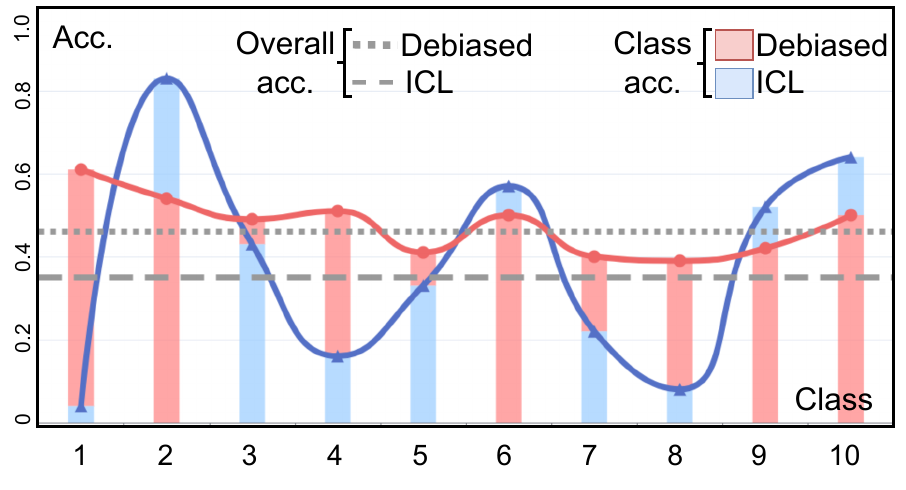}
  \end{center}
  \caption{The accuracy imbalance and what we can achieve with debiasing}
  \label{fig:biasissue}
\end{wrapfigure}

Language models are good few-shot learners, but prompting can result in inevitable class accuracy imbalance. A language model can perform significantly better on certain classes compared to others, leading to biased performances across different classes (illustrated in Figure \ref{fig:biasissue}). This issue is particularly relevant in tasks like classification, or any scenario where the model must distinguish between multiple categories. For example, on DBpedia ontology classification \citep{auer2007}, the overall accuracy for 14 classes can reach 88\% when prompting Llama-2-13B \citep{llama2}, but class \textit{Nature}'s accuracy is 28\%. Broadly, for open-ended generation tasks, when treating each token in the vocabulary as a distinct class, generating the most likely next token can be alternatively viewed as classifying over the entire vocabulary or soft labels \citep{Thrampoulidis1,Thrampoulidis2}, also susceptible to class accuracy imbalance. Therefore, accuracy imbalance is prevalent and affecting users of language models, where classification is the most basic case manifesting such imbalances.

In text classifications, the prevalence of class accuracy imbalance is sometimes hidden by a single or a dominant evaluation metric, such as overall accuracy. Especially for relatively large models, the fact that they excel on benchmarks which emphasize overall performances may hide a poor performance in some classes. To enable a focused evaluation of whether language model performances are balanced across classes, the COBias metric was proposed \citep{cobias} to dedicate evaluating pairwise class accuracy differences.

The class accuracy imbalance issue is not particularly easy to solve from the root. Its causes come from within the model architecture. In details, language models rely on priors learned during training to make predictions. However, built on self-attention, the models are prone to the copying mechanism of the induction heads \citep{elhage2021,olsson2022}. These heads predict repeated sequences, which can over-capture surface-level patterns in the data, leading to biased priors, and then biased predictions. Therefore, if certain classes or concepts were more frequently emphasized in the training data, the model may disproportionately favor them \citep{zhao2021}. Although a systemic change at model level or selecting high-quality data may help, it can be inefficient.

Besides what is rooted in the model and data, the way prompts are phrased can further bias the model towards certain classes \citep{framing,clinical,sclar2024}. In few-shot prompting, imbalanced class accuracy can be exacerbated because of underrepresented classes in the demonstrations of a prompt. Nevertheless, it only helps to a limited extent by making all classes well-represented in the demonstrations \citep{cobias}. In addition, Chain-of-Thought style outputs may facilitate self-correcting responses without relying on prompt engineering \citep{denny2024}, but the outcome is guided and restricted by reward signals. Lastly, although certain types of biases, such as gender bias or truthfulness, can be traced back to Transformer layers and intervened at inference time \citep{Vig2020,Geiger2021,kenneth2023}, interventions can be inefficient to mitigate the accuracy bias, especially when it could be better treated without altering of the model's internal structure.

Therefore, to effectively and efficiently reduce class accuracy imbalances, post-hoc optimization methods are introduced. As such, post-hoc probability correction complements training and prompting strategies to prevent the model from overly relying on learned patterns, and to incorporate necessary information in downstream tasks. This type of correction involves correcting In-context learned (ICL) output probabilities in inference time by combinatorially optimizing a relatively small mathematical model. These direct rectifications are useful, because they can directly correct outputs, no matter what biases are encoded in the model prior, or what prompts are used.

\noindent \textbf{Related Work.} Prior post-hoc correction work has investigated re-weighting ICL output probabilities through combinatorial optimization addressing COBias\citep{cobias,furud} or calibration addressing biased predictions \citep{bc,zhao2021}. The former uses labeled instances for optimization while the latter does not. Both lines of work advance overall classification accuracy with low costs and the flexibility of no language model parameter updates. The performance difference is that the combinatorial optimization works achieve better balanced class performances while improving overall accuracy. We follow the combinatorial optimization perspective for our ensemble debiasing.

For a background on combinatorial optimization, it provides a mathematical framework for maximizing or minimizing an objective function of decision variables, subject to constraints on some or all of the variables, involving inequalities, equalities, and integrality restrictions \citep{nemhauser1988}. This framework essentially transforms a rich variety of tasks into discrete optimization models, which is robust and remarkably versatile. It solves real-world problems such as planning, scheduling, etc. in operations research, and has diverse applications in NLP, from structured prediction such as named entity recognition and dependency parsing, to reading comprehension, then to grammatical error correction, and to align LLM outputs with human preferences \citep{martins2009,wang2010,rush2010,koo2010,goldwasser2012,wu2013,berant2014,roth2017,lin2021,ner4opt,zhang2024human,srikumar2023,garmendia2024survey}. The core use of the framework is to translate a problem into a mathematical statement of one of the forms of linear mixed-integer programming problem (MIP), integer programming problem (IP), nonlinear integer programming problem (NIP), or other types \citep{roth2004,roth2005,hemmecke2010,srikumar2023}. Solving combinatorial optimization models is not an easy task. Among which, NIP is highly challenging, because the NIP model can contain discontinuous functions and is inherently non-differentiable. \citet{cobias,furud} proposed to address the search difficulty with metaheuristics, which we follow (more discussions in Appendix \ref{appdix:bnb}).

Back to the combinatorial optimization based debiasing methods, their drawbacks lie in the single consideration at either class level \citep{cobias} or sample level \citep{furud}. In \citep{cobias}, any sample gets re-weighted by the same set of per-class correction weights, making it hard to capture sample-level variances, which is particularly needed for weak classes. In \citep{furud}, corrections are made at sample level by assigning per-sample per-class correction weights that are computed by membership functions, however, capturing too much nuance may over-corrects strong classes.

To overcome the insufficiency in previous works, we propose a meticulous post-hoc correction framework \textbf{DCS} (\textbf{D}ebiasing at \textbf{C}lass and \textbf{S}ample levels). DCS is an organic integration of broad and fine-grained corrections across class and sample levels, pushing forward fairer prompting accuracy for LLMs. DCS automatically detects which classes need corrections that account for sample-level differences and which do not. In the following, we first introduce the Heaviside step function based ensembling strategy for DCS (Section \ref{sec:model}) and the overall framework (Section \ref{sec:sa}), followed by quantitative evaluations and in-depth analyses (Section \ref{sec:exp}). In Section \ref{sec:concl}, we conclude with prospective applications and open problems.

Our contributions are summarized as below:

\begin{itemize}
    \item \textbf{\textit{Post-hoc debiasing unifies class and sample level corrections.}} We ensemble weight correction and membership correction to enable rectifications of class probabilities at both class and sample levels, to enhance the performance of LLMs directly from their outputs.

    \item \textbf{\textit{SoTA scores on overall accuracy while maintaining low COBias}} DCS can achieve leading overall accuracy improvements while good COBias reduction across seven benchmark text classification tasks compared to the previous SoTA. 
    
    \item \textbf{\textit{Optimal solutions need weight correction to elevate weak classes.}} Quantitative analyses show that the average ratio of classes using membership correction to weight correction is between 0 and 1, where most classes apply weight corrections but lowest-accuracy (weakest) classes more often apply membership corrections, suggesting the necessity of keeping membership correction as an option and letting the optimization process determine if it is needed for any particular class.
\end{itemize}

\section{Using Heaviside Step Function to Ensemble Class and Sample Level Debiasing}
\label{sec:model}
We introduce a robust ensemble debiasing framework across class and sample levels, which is targeted specifically for balancing class accuracy while improving overall accuracy over individual methods. To achieve ensembling, a Heaviside step function seamlessly unifies two unique ICL output correction methods, bringing together the advantages of weight coefficient and fuzzy rule based corrections. This unified framework captures the intricacies of ICL output class probabilities with greater precision, reaching solutions that elevate weak classes while maintaining the strong ones. List of notations are presented in Appendix \ref{appdix:notations}.

\subsection{The Mapping Function for ICL Probability Correction}
As mentioned, a common method for correcting probabilities involves multiplying each class probability by a coefficient \citep{cobias}. This is a class-level correction. For an $N$-class task, we use a set of weights to adjust each dimension of the ICL output $N$-dimensional probabilities without considering the variations in the initial probabilities across different instances. In contrast, a more refined correction method employs membership functions from fuzzy logic \citep{furud}, where a mapping function is applied to adjust each dimension. Different classes use different mapping functions. As a result, class probabilities of different instances undergo varying degrees of transformation - some are adjusted significantly, others minimally, and some remain unchanged, reflecting a finer, sample-level correction approach.

We propose enabling a model-driven selection between coarse (class-level) and finer (sample-level) corrections in the optimization process, promoting efficient and precise post-hoc corrections for any class. Let's denote the probability of the $m$-th instance's $i$-th class as $p_{mi}$. We correct the per-sample per-class probability by a mapping function $f_i: p'_{mi} \leftarrow f_i(p_{mi}), p_{mi} \in [0,1]$. To achieve our goal, the set of mapping functions $\Psi$ combine weight correction with membership function correction, i.e., $\Psi = \mathbb{R}\textsuperscript{Weight} \cup \mathbb{R}\textsuperscript{Membership}$, where $\mathbb{R}\textsuperscript{Weight}=\{\omega_1,\dots,\omega_{D_W}\}$ and $\mathbb{R}\textsuperscript{Membership}=\{\mu_1,\dots,\mu_{D_F}\}$. To dynamically choose the correction, we first introduce the Heaviside step function:
\begin{equation}
        H(x)= 
\begin{cases}
    1,& x\geq 0\\
    0,              & \text{otherwise}
\end{cases}
\label{eq:1}
\end{equation}
Let $\xi_i$ denote the selection variable for class $i$, each $\xi_i$ will be optimized to choose a most suitable correction from $\Psi$. We conventionally note when $\xi_i$ is optimized to be one of the indices from $\{1,...,D_F\}$, the i-th class probability of each instance gets a personalized adjustment by a membership function determined by $\xi_i$; when $\xi_i$ is one of the indices from $\{D_F+1,...,D_F+D_W\}$, the i-th class probability of any input instance is corrected by a weight coefficient determined by $\xi_i$. Consequently, when $\xi_i=k$ and $k \le D_F$, the triangular membership function (following \citep{furud}) based correction function $\mu_k$ is selected, the updated probability is computed by:
\begin{equation}
   \mu_k(p_{mi}) = \left \{
  \begin{aligned}
    &0, && p_{mi} \le a_k\\
    &\frac{p_{mi}-a_k}{b_k-a_k}, && a_k < p_{mi} \le b_k\\
    &\frac{c_k-p_{mi}}{c_k-b_k}, && b_k < p_{mi} \le c_k\\
    &0, && \text{otherwise}
  \end{aligned} \right.
  \label{eq:2}   
\end{equation}
Special cases: when $a_k=b_k=0, \mu_k(p_{mi})=\frac{c_k-p_{mi}}{c_k}$ for $0 \le p_{mi} \le c_k$ and 0 for $p_{mi} \ge c_k$; when $b_k=c_k=1, \mu_k(p_{mi})=\frac{p_{mi}-a_k}{1-a_k}$ for $a_k \le p_{mi} \le 1$ and 0 for $p_{mi} \le a_k$.

Otherwise, the weight correction function $\omega_k$ is selected, and the updated probability is obtained by:
\begin{equation}
\omega_k(p_{mi}) = \frac{k-D_F}{D_W}p_{mi}
  \label{eq:3}   
\end{equation}
Therefore, the per-sample per-class probability $p_{mi}$ is precisely updated as follows.
\begin{equation}
    p'_{mi}=f_{i}(p_{mi})=\mu_{\xi_i}(p_{mi}) \cdot H(D_F-\xi_i)+ \omega_{\xi_i}(p_{mi}) \cdot H(\xi_i-D_F-1)
    \label{eq:4}
\end{equation}
This results in a more accurate correction for the ICL output probability, due to applying flexible fixes at either the broad level or the fine-grained level. 

\subsection{Mathematical Model}
The mathematical model used to drive the selection is based on the nonlinear integer programming model in \citep{cobias} with accuracy-maximizing and COBias-minimizing objectives. The combination of weight and membership corrections that optimizes the model's objective function will be selected. In short, the model is:
\begin{align}
    \min Z &= Z^{\text{Err}}(\xi)+\beta Z^{\text{COBias}}(\xi)+\tau Z^{\text{PMI}}(\xi) \nonumber \\ 
    \text{s.t.   } & \boldsymbol{\xi}=(\xi_1,\dots,\xi_N), \xi_i \in \{1,\dots,D_F+D_W\}
\end{align}
where $Z^{\text{Err}}, Z^{\text{COBias}},Z^{\text{PMI}}$ respectively corresponds to the overall error rate, COBias of all class pairs, and point-wise mutual information (PMI) between predicted and ground-truth classes (Appendix \ref{appdix:obj}). During optimization, at the beginning of each simulated annealing iteration, candidate selection $\xi_i$'s are generated and the correction functions $f_i$'s correct each sample in the optimization set, yielding corrected predictions $\{\hat{y}_m\}_{m=1}^M$. Each of the individual term in $Z$ is essentially an evaluation score computed using the corrected predictions and ground-truth labels $\{y_m\}_{m=1}^M$. If the new $Z$ is smaller than the current $Z$, the candidate selection is accepted and the best selection is updated; otherwise, the algorithm decides whether to accept the new selection based on the acceptance probability. Iterations repeat until a stopping criterion is met. The optimization process will be elaborated in Section \ref{sec:sa}.

\noindent \textbf{Model parameters.} This is a budget-friendly model with a total number of parameters (here we refer to search space size) of $N(D_F+D_W)$.

\section{Simulated Annealing for Solving DCS}
\label{sec:sa}
Due to the non-differentiable nature of the math model, we turn to heuristics and employ the simulated annealing (SA) algorithm to solve the DCS model. The SA algorithm was proposed by \citep{kirkpatrick1983,Cerny}. The forerunners successfully introduced the concept of annealing into combinatorial optimization, inspired by the annealing process of solid materials. It starts at an initial temperature and cools down at a specified rate. As the temperature gradually decreases, the algorithm leverages probabilistic jumps to explore the solution space randomly, searching for the global optimum of the objective function, so it can probabilistically escape local optima and eventually converge to the global optimum \citep{Steinbrunn}. The optimization of DCS model is a search problem in multidimensional space, where each solution makes a potential \textit{\textbf{probability correction scheme}}. Next, we uncover the overall framework that apply SA to solve our model.

\subsection{Overall Framework}
\begin{figure}
    \centering
    \includegraphics[width=0.95\linewidth]{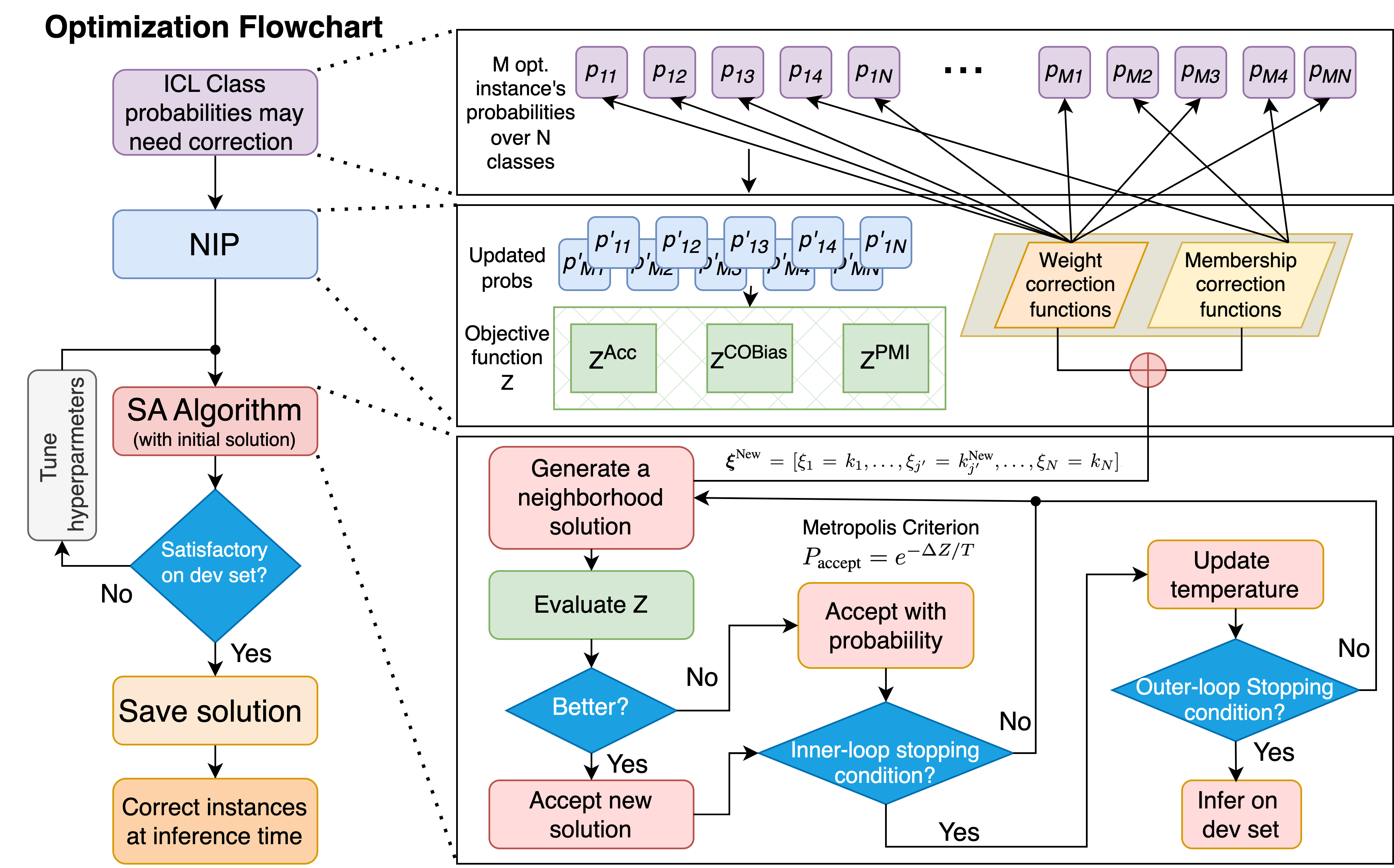}
    \caption{Overall framework of DCS}
    \label{fig:overall}
\end{figure}

The overall solution framework for DCS is depicted in Fig. \ref{fig:overall}. The combinatorial optimization relies on a labeled set of these initial class probabilities and their ground-truth labels. The labeled set can be simply taken from the full or a subset of the classification task's training set. After ICL output class probabilities are obtained, the optimization process starts with an initial solution. The iteration repeats until either a minimum temperature is reached or a maximum number of iterations is reached, and we obtain the optimal correction functions. In actual inference, the selected functions directly rectify a test instance's ICL output class probabilities and result in corrected predictions.

\subsection{Initial Solution}
We can keep the initial probabilities unchanged as the initial solution for the DCS math model. The fuzzy rule set $\mathbb{R}\textsuperscript{Membership}=\{\mu_k\}_{k=1}^{D_F}$ designed in this paper must include \textit{Don't Change}, the triangular membership function with $a_k=0, b_k=c_k=1$, which will keep the original probability unchanged. If the corresponding selection index is $k_0$, the initial solution can be chosen as: $\xi_j = k_0, \text{ for class } j=1,\dots,N$. For SA, the initial solution does not affect the quality of the final solution, but it influences the annealing process, and we empirically start with unchanged probabilities \citep{kirkpatrick1983}. 

\subsection{Neighborhood Solution}
It is necessary to generate a new solution in the neighborhood of the current solution. If the current solution is $\xi_j = k_j, \text{ for class } j=1,\dots,N$, we can randomly select a class, for example, \(j'\) whose current solution is $\xi_{j'} = k_{j'}$, and perturb it by sampling $k_{j'}^{\text{New}}$ from the set $\mathbb{D}\setminus k_{j'}$ as the new value for the variable $\xi_{j'}$. That is, $\xi_{j'} = k_{j'}^{\text{New}}$. In this way, we obtain a neighboring solution: $\boldsymbol{\xi}^{\text{New}} =(\xi_1=k_1, \dots, \xi_{j'}=k_{j'}^{\text{New}}, \dots, \xi_N=k_N)$. The transition from the current solution at a given temperature to the new solution is a Markov chain \citep{kirkpatrick1983,Cerny}, and the transition probability is determined by the Metropolis criterion $P_{\text{accept}}=e^{-\Delta Z/T}$. If the new solution is better ($\Delta Z<0$), accept the new solution; if the new solution is worse ($\Delta Z \ge 0$) and a uniform random number $r \sim U(0,1)$ is smaller than $P_{\text{accept}}$, namely $r < P_{\text{accept}}$, the new solution is also accepted.

\subsection{Annealing Schedule}
An annealing schedule is characterized by the initial value of the temperature, the cooling rate, the inner-loop criterion, and the outer-loop criterion.  

\noindent \textbf{Initial Temperature:}The initial temperature is chosen so that the acceptance probability is relatively close to one. The higher the initial temperature, the more possible to obtain a high-quality solution, but the longer the computation time. In this work, the initial temperature is set to 200,000.  

\noindent \textbf{Inner-Loop Criterion:} The inner loop simulates the process of the system reaching equilibrium at a given temperature. Theoretically, there should be a sufficient number of iterations. The more decision variables there are, the more iterations the inner loop should have. Therefore, in this paper, the inner loop stops when either the number of accepted solutions reaches $\lambda_1$ times of the number of variables, or the number of solutions generated exceeds $\lambda_2$ times of the number of variables. The value of $\lambda_1, \lambda_2$ are tuned alongside other hyperparameters. 

\noindent \textbf{Outer-Loop Criterion and Cooling Rate:} Generally, the termination temperature should be set to a sufficiently small positive number. The temperature update function follows a geometric rate:  $T_{\text{curr}+1} = \alpha \cdot T_{\text{curr}}$, where $\alpha$ is set as 0.95 in this paper.

\section{Experiments}
\label{sec:exp}
We assess the capabilities of DCS in reducing class accuracy differences and enhancing overall accuracy with three ICL settings: 1-shot, 5-shot, and $N$-shot. Demonstrations in the prompt are randomly selected from the optimization set for 1-shot and 5-shot cases, while the $N$-shot demonstration cascades examples from each class, to minimize under-representations of any class. DCS is compared against SoTA methods, DNIP \citep{cobias} and FuRud \citep{furud}. We prompt LLMs with an A100 GPU to obtain initial ICL outputs, following DNIP's hyperparameters, then we run simulated annealing on CPUs. Evaluations are performed on seven multi-class text classification tasks across general and biomedical domains, including AGNews (AGN, 4-class;\citep{zhang2015}), DBpedia (DBP, 14-class;\citep{auer2007}), SST-5 (SST, 5-class;\citep{socher2013}), TREC (6-class;\citep{voorhees2000,li2002}), RTE (binary;\citep{dagan2006}), DDI (5-class;\citep{ddi}), and PubMedQA (PMQA, 3-class;\citep{pubmedqa}). Appendix \ref{appdix:setup} presents more details.

\subsection{Main Results}
\noindent \textbf{Setup} We perform quantitative evaluations on Llama-2-13B, for its representativeness as a widely applied open-source LLM. Indeed, Llama-2-13B is not particularly big in parameter size, but its architecture can be seen as basic building blocks for many advanced LLMs \citep{vicuna,codallama,llava}, and we expect similar findings in larger model variants or models similar in architecture, which will be discussed in Section \ref{subsec:largermodel}. Table \ref{tab:main} reports our main findings.

\begin{table}[ht]
\centering
\resizebox{\textwidth}{!}{%
\begingroup
    \renewcommand{\arraystretch}{2}
\begin{tabular}{@{}lcccccccccccccccccccc@{}}
\toprule
\multicolumn{1}{l|}{} & \multicolumn{10}{c|}{\Huge Acc} & \multicolumn{10}{c}{\Huge COBias} \\ \cmidrule(l){2-21} 
\multicolumn{1}{l|}{} & \multicolumn{1}{c|}{} & \multicolumn{6}{c|}{\LARGE General Domain} & \multicolumn{3}{c|}{\LARGE Biomedical Domain} & \multicolumn{1}{c|}{} & \multicolumn{6}{c|}{\LARGE General Domain} & \multicolumn{3}{c}{\LARGE Biomedical Domain} \\ \cmidrule(lr){3-11} \cmidrule(l){13-21} 
\multicolumn{1}{l|}{\multirow{-3}{*}{\Huge Method}} & \multicolumn{1}{c|}{\multirow{-2}{*}{\LARGE All}} & \multicolumn{1}{c|}{\LARGE Avg} & \LARGE  AGN & \LARGE DBP & \LARGE SST & \LARGE TREC & \multicolumn{1}{c|}{\LARGE RTE} & \multicolumn{1}{c|}{\LARGE Avg} & \LARGE DDI & \multicolumn{1}{c|}{\LARGE PMQA} & \multicolumn{1}{c|}{\multirow{-2}{*}{\LARGE All}} & \multicolumn{1}{c|}{\LARGE Avg} & \LARGE AGN & \LARGE DBP & \LARGE SST & \LARGE TREC & \multicolumn{1}{c|}{\LARGE RTE} & \multicolumn{1}{c|}{\LARGE \LARGE Avg} & \LARGE DDI & \LARGE PMQA \\ \midrule
\multicolumn{21}{c}{\LARGE \textit{1-shot ICL + post-hoc debiasing}} \\ \midrule
\multicolumn{1}{l|}{\LARGE ICL} & \multicolumn{1}{c|}{\LARGE 59.4} & \multicolumn{1}{c|}{\LARGE 70.7} & \LARGE 79.9\textsubscript{7.0} & \LARGE 88.2\textsubscript{1.0} & \LARGE 44.9\textsubscript{4.3} & \LARGE 68.5\textsubscript{10.8} & \multicolumn{1}{c|}{\LARGE 71.5\textsubscript{2.2}} & \multicolumn{1}{c|}{\LARGE 31.2} & \LARGE 7.2\textsubscript{0.9} & \multicolumn{1}{c|}{\LARGE 55.1\textsubscript{2.9}} & \multicolumn{1}{c|}{\LARGE 40.5} & \multicolumn{1}{c|}{\LARGE 35.4} & \LARGE 28.3\textsubscript{16.1} & \LARGE 16.2\textsubscript{3.7} & \LARGE 53.1\textsubscript{5.0} & \LARGE 35.9\textsubscript{6.5} & \multicolumn{1}{c|}{\LARGE 43.4\textsubscript{7.0}} & \multicolumn{1}{c|}{\LARGE 53.4} & \LARGE 45.6\textsubscript{5.9} & \LARGE 61.2\textsubscript{1.9} \\
\multicolumn{1}{l|}{\LARGE +DNIP} & \multicolumn{1}{c|}{\LARGE 69.2} & \multicolumn{1}{c|}{\LARGE 76.2} & \LARGE 87.9\textsubscript{0.7} & \LARGE 93.4\textsubscript{0.6} & \LARGE 48.3\textsubscript{1.9} & \LARGE 77.1\textsubscript{2.0} & \multicolumn{1}{c|}{\LARGE 74.3\textsubscript{0.8}} & \multicolumn{1}{c|}{\LARGE 51.8} & \LARGE 40.4\textsubscript{6.0} & \multicolumn{1}{c|}{\LARGE 63.1\textsubscript{14.0}} & \multicolumn{1}{c|}{\cellcolor[HTML]{F9EAEA}\LARGE 14.3} & \multicolumn{1}{c|}{\cellcolor[HTML]{F9EAEA}\LARGE 10.2} & \LARGE 6.3\textsubscript{0.6} & \LARGE 7.7\textsubscript{0.6} & \cellcolor[HTML]{F9EAEA}\LARGE 18.7\textsubscript{10.1} & \LARGE 14.2\textsubscript{1.3} & \multicolumn{1}{c|}{\cellcolor[HTML]{F9EAEA}\LARGE 4.3\textsubscript{3.3}} & \multicolumn{1}{c|}{\LARGE 24.3} & \cellcolor[HTML]{F9EAEA}\LARGE 7.5\textsubscript{3.2} & \LARGE 41.1\textsubscript{29.6} \\
\multicolumn{1}{l|}{\LARGE  +FuRud} & \multicolumn{1}{c|}{\LARGE 67.2} & \multicolumn{1}{c|}{\LARGE 76.5} & \LARGE 86.8\textsubscript{2.0} & \LARGE 92.7\textsubscript{0.2} & \LARGE 50.0\textsubscript{3.0} & \LARGE 78.4\textsubscript{2.4} & \multicolumn{1}{c|}{\cellcolor[HTML]{F9EAEA}\LARGE 74.5\textsubscript{1.8}} & \multicolumn{1}{c|}{\LARGE 44.0} & \LARGE 29.4\textsubscript{17.0} & \multicolumn{1}{c|}{\LARGE 58.6\textsubscript{4.1}} & \multicolumn{1}{c|}{\LARGE 15.1} & \multicolumn{1}{c|}{\LARGE 13.3} & \LARGE 10.9\textsubscript{4.7} & \LARGE 8.8\textsubscript{0.3} & \LARGE 26.2\textsubscript{11.6} & \LARGE 13.5\textsubscript{1.4} & \multicolumn{1}{c|}{\LARGE 7.1\textsubscript{1.7}} & \multicolumn{1}{c|}{\cellcolor[HTML]{F9EAEA}\LARGE 19.5} & \LARGE 11.8\textsubscript{6.7} & \cellcolor[HTML]{F9EAEA}\LARGE 27.2\textsubscript{8.1} \\
\multicolumn{1}{l|}{\LARGE  +DCS (Ours)} & \multicolumn{1}{c|}{\cellcolor[HTML]{F9EAEA}\LARGE 72.2} & \multicolumn{1}{c|}{\cellcolor[HTML]{F9EAEA}\LARGE 77.8} & \cellcolor[HTML]{F9EAEA}\LARGE 88.2\textsubscript{1.0} & \cellcolor[HTML]{F9EAEA}\LARGE 94.7\textsubscript{0.7} & \cellcolor[HTML]{F9EAEA}\LARGE 50.2\textsubscript{1.4} & \cellcolor[HTML]{F9EAEA}\LARGE 81.5\textsubscript{2.5} & \multicolumn{1}{c|}{\LARGE 74.4\textsubscript{0.4}} & \multicolumn{1}{c|}{\cellcolor[HTML]{F9EAEA}\LARGE 58.1} & \cellcolor[HTML]{F9EAEA}\LARGE 52.9\textsubscript{18.3} & \multicolumn{1}{c|}{\cellcolor[HTML]{F9EAEA}\LARGE 63.3\textsubscript{12.2}} & \multicolumn{1}{c|}{\LARGE 15.0} & \multicolumn{1}{c|}{\LARGE 10.7} & \cellcolor[HTML]{F9EAEA}\LARGE 6.2\textsubscript{1.5} & \cellcolor[HTML]{F9EAEA}\LARGE 6.0\textsubscript{0.8} & \LARGE 23.7\textsubscript{11.6} & \cellcolor[HTML]{F9EAEA}\LARGE 12.8\textsubscript{3.3} & \multicolumn{1}{c|}{\LARGE 5.0\textsubscript{3.6}} & \multicolumn{1}{r|}{\LARGE 25.8} & \LARGE 16.4\textsubscript{13.5} & \LARGE 35.2\textsubscript{22.3} \\ \midrule
\multicolumn{21}{c}{\LARGE \textit{5-shot ICL + post-hoc debiasing}} \\ \midrule
\multicolumn{1}{l|}{\LARGE ICL} & \multicolumn{1}{c|}{\LARGE 63.9} & \multicolumn{1}{c|}{\LARGE 68.5} & \LARGE 82.5\textsubscript{2.0} & \LARGE 93.6\textsubscript{1.3} & \LARGE 45.8\textsubscript{6.0} & \LARGE 58.7\textsubscript{23.3} & \multicolumn{1}{c|}{\LARGE 61.9\textsubscript{16.9}} & \multicolumn{1}{c|}{\LARGE 52.3} & \LARGE 34.1\textsubscript{42.1} & \multicolumn{1}{c|}{\LARGE 70.4\textsubscript{7.1}} & \multicolumn{1}{c|}{\LARGE 39.2} & \multicolumn{1}{c|}{\LARGE 35.6} & \LARGE 24.4\textsubscript{4.2} & \LARGE 9.0\textsubscript{2.0} & \LARGE 48.0\textsubscript{15.4} & \LARGE 35.4\textsubscript{13.7} & \multicolumn{1}{c|}{\LARGE 61.3\textsubscript{40.4}} & \multicolumn{1}{c|}{\LARGE 48.3} & \LARGE 44.4\textsubscript{5.4} & \LARGE 52.2\textsubscript{19.3} \\
\multicolumn{1}{l|}{\LARGE  +DNIP} & \multicolumn{1}{c|}{\LARGE 71.9} & \multicolumn{1}{c|}{\LARGE 77.9} & \LARGE 88.5\textsubscript{0.6} & \LARGE 95.8\textsubscript{0.7} & \LARGE 52.9\textsubscript{2.5} & \LARGE 76.6\textsubscript{8.7} & \multicolumn{1}{c|}{\LARGE 75.8\textsubscript{4.5}} & \multicolumn{1}{c|}{\LARGE 57.0} & \LARGE 54.1\textsubscript{5.1} & \multicolumn{1}{c|}{\LARGE 59.9\textsubscript{1.4}} & \multicolumn{1}{c|}{\cellcolor[HTML]{ECF4FF}\LARGE 10.3} & \multicolumn{1}{c|}{\LARGE 9.2} & \LARGE 7.0\textsubscript{0.9} & \LARGE 5.7\textsubscript{1.1} & \LARGE 15.8\textsubscript{12.6} & \LARGE 14.4\textsubscript{7.5} & \multicolumn{1}{c|}{\LARGE 3.1\textsubscript{1.9}} & \multicolumn{1}{c|}{\cellcolor[HTML]{ECF4FF}\LARGE 13.1} & \cellcolor[HTML]{ECF4FF}\LARGE 16.5\textsubscript{7.6} & \cellcolor[HTML]{ECF4FF}\LARGE 9.6\textsubscript{4.6} \\
\multicolumn{1}{l|}{\LARGE  +FuRud} & \multicolumn{1}{c|}{\LARGE 70.6} & \multicolumn{1}{c|}{\LARGE 77.9} & \LARGE 87.8\textsubscript{0.9} & \LARGE 95.5\textsubscript{0.7} & \LARGE 53.3\textsubscript{4.9} & \LARGE 76.9\textsubscript{3.9} & \multicolumn{1}{c|}{\LARGE 75.8\textsubscript{4.2}} & \multicolumn{1}{c|}{\LARGE 52.3} & \LARGE 40.3\textsubscript{7.3} & \multicolumn{1}{c|}{\LARGE 64.3\textsubscript{5.6}} & \multicolumn{1}{c|}{\LARGE 14.7} & \multicolumn{1}{c|}{\LARGE 11.8} & \LARGE 10.6\textsubscript{0.7} & \LARGE 5.7\textsubscript{0.9} & \LARGE 22.1\textsubscript{6.0} & \LARGE 19.6\textsubscript{10.7} & \multicolumn{1}{c|}{\cellcolor[HTML]{ECF4FF}\LARGE 1.0\textsubscript{0.3}} & \multicolumn{1}{c|}{\LARGE 22.1} & \LARGE 19.1\textsubscript{3.9} & \LARGE 25.1\textsubscript{19.5} \\
\multicolumn{1}{l|}{\LARGE  +DCS (Ours)} & \multicolumn{1}{c|}{\cellcolor[HTML]{ECF4FF}\LARGE 74.8} & \multicolumn{1}{c|}{\cellcolor[HTML]{ECF4FF}\LARGE 79.0} & \cellcolor[HTML]{ECF4FF}\LARGE 88.8\textsubscript{0.8} & \cellcolor[HTML]{ECF4FF}\LARGE 96.3\textsubscript{0.6} & \cellcolor[HTML]{ECF4FF}\LARGE 54.2\textsubscript{2.9} & \cellcolor[HTML]{ECF4FF}\LARGE 79.7\textsubscript{4.9} & \multicolumn{1}{c|}{\cellcolor[HTML]{ECF4FF}\LARGE 76.1\textsubscript{4.3}} & \multicolumn{1}{c|}{\cellcolor[HTML]{ECF4FF}\LARGE 64.3} & \cellcolor[HTML]{ECF4FF}\LARGE 57.1\textsubscript{4.2} & \multicolumn{1}{c|}{\cellcolor[HTML]{ECF4FF}\LARGE 71.4\textsubscript{8.5}} & \multicolumn{1}{c|}{\LARGE 14.7} & \multicolumn{1}{c|}{\cellcolor[HTML]{ECF4FF}\LARGE 7.7} & \cellcolor[HTML]{ECF4FF}\LARGE 6.7\textsubscript{0.7} & \cellcolor[HTML]{ECF4FF}\LARGE 4.7\textsubscript{1.1} & \cellcolor[HTML]{ECF4FF}\LARGE 11.7\textsubscript{8.3} & \cellcolor[HTML]{ECF4FF}\LARGE 14.0\textsubscript{4.1} & \multicolumn{1}{c|}{\LARGE 1.6\textsubscript{0.6}} & \multicolumn{1}{c|}{\LARGE 32.2} & \LARGE 21.0\textsubscript{4.4} & \LARGE 43.4\textsubscript{23.4} \\ \midrule
\multicolumn{21}{c}{\LARGE  \textit{N-shot ICL + post-hoc debiasing}} \\ \midrule
\multicolumn{1}{l|}{\LARGE ICL} & \multicolumn{1}{c|}{\LARGE 61.9} & \multicolumn{1}{c|}{\LARGE 74.2} & \LARGE 83.5\textsubscript{1.5} & \LARGE 95.2\textsubscript{1.2} & \LARGE 50.3\textsubscript{2.3} & \LARGE 67.0\textsubscript{12.7} & \multicolumn{1}{c|}{\LARGE 75.0\textsubscript{0.8}} & \multicolumn{1}{c|}{\LARGE 31.0} & \LARGE 9.7\textsubscript{1.0} & \multicolumn{1}{c|}{\LARGE 52.3\textsubscript{5.3}} & \multicolumn{1}{c|}{\LARGE 25.6} & \multicolumn{1}{c|}{\LARGE 23.8} & \LARGE 14.9\textsubscript{5.1} & \LARGE 7.0\textsubscript{2.2} & \LARGE 36.3\textsubscript{7.2} & \LARGE 38.2\textsubscript{5.1} & \multicolumn{1}{c|}{\LARGE 22.5\textsubscript{13.2}} & \multicolumn{1}{c|}{\LARGE 30.3} & \LARGE 39.7\textsubscript{3.5} & \LARGE 20.9\textsubscript{4.2} \\
\multicolumn{1}{l|}{\LARGE  +DNIP} & \multicolumn{1}{c|}{\LARGE 70.7} & \multicolumn{1}{c|}{\LARGE 79.2} & \LARGE 88.7\textsubscript{0.5} & \LARGE 96.6\textsubscript{0.5} & \LARGE 51.3\textsubscript{1.0} & \LARGE 82.7\textsubscript{1.4} & \multicolumn{1}{c|}{\LARGE 76.7\textsubscript{3.7}} & \multicolumn{1}{c|}{\LARGE 49.5} & \LARGE 43.6\textsubscript{6.1} & \multicolumn{1}{c|}{\LARGE 55.3\textsubscript{2.6}} & \multicolumn{1}{c|}{\cellcolor[HTML]{FCFBD8}\LARGE 7.5} & \multicolumn{1}{c|}{\cellcolor[HTML]{FCFBD8}\LARGE 6.3} & \LARGE 7.3\textsubscript{0.5} & \LARGE 4.3\textsubscript{0.7} & \cellcolor[HTML]{FCFBD8}\LARGE 2.8\textsubscript{1.5} & \LARGE 12.1\textsubscript{5.5} & \multicolumn{1}{c|}{\cellcolor[HTML]{FCFBD8}\LARGE 5.0\textsubscript{3.3}} & \multicolumn{1}{c|}{\cellcolor[HTML]{FCFBD8}\LARGE 10.6} & \cellcolor[HTML]{FCFBD8}\LARGE 12.5\textsubscript{4.3} & \cellcolor[HTML]{FCFBD8}\LARGE 8.7\textsubscript{1.2} \\
\multicolumn{1}{l|}{\LARGE  +FuRud} & \multicolumn{1}{c|}{\LARGE 73.6} & \multicolumn{1}{c|}{\LARGE 78.7} & \LARGE 87.9\textsubscript{0.8} & \LARGE 96.5\textsubscript{0.5} & \LARGE 53.7\textsubscript{2.5} & \LARGE 78.3\textsubscript{5.0} & \multicolumn{1}{c|}{\cellcolor[HTML]{FFFFC7}\LARGE 77.1\textsubscript{3.2}} & \multicolumn{1}{c|}{\LARGE 60.7} & \LARGE 61.7\textsubscript{6.8} & \multicolumn{1}{c|}{\LARGE 59.7\textsubscript{7.2}} & \multicolumn{1}{c|}{\LARGE 16.2} & \multicolumn{1}{c|}{\LARGE 13.1} & \LARGE 8.5\textsubscript{3.2} & \LARGE 4.9\textsubscript{0.7} & \LARGE 29.0\textsubscript{11.6} & \LARGE 15.8\textsubscript{3.1} & \multicolumn{1}{c|}{\LARGE 7.3\textsubscript{4.9}} & \multicolumn{1}{c|}{\LARGE 23.9} & \LARGE 28.9\textsubscript{3.8} & \LARGE 18.8\textsubscript{14.9} \\
\multicolumn{1}{l|}{\LARGE  +DCS (Ours)} & \multicolumn{1}{c|}{\cellcolor[HTML]{FCFBD8}\LARGE 74.7} & \multicolumn{1}{c|}{\cellcolor[HTML]{FCFBD8}\LARGE 80.0} & \cellcolor[HTML]{FCFBD8}\LARGE 88.9\textsubscript{0.5} & \cellcolor[HTML]{FCFBD8}\LARGE 96.8\textsubscript{0.4} & \cellcolor[HTML]{FCFBD8}\LARGE 54.2\textsubscript{0.6} & \cellcolor[HTML]{FCFBD8}\LARGE 83.4\textsubscript{3.1} & \multicolumn{1}{c|}{\LARGE 76.7\textsubscript{3.7}} & \multicolumn{1}{c|}{\cellcolor[HTML]{FCFBD8}\LARGE 61.4} & \cellcolor[HTML]{FCFBD8}\LARGE 62.4\textsubscript{16.5} & \multicolumn{1}{c|}{\cellcolor[HTML]{FCFBD8}\LARGE 60.3\textsubscript{5.3}} & \multicolumn{1}{c|}{\LARGE 13.0} & \multicolumn{1}{c|}{\LARGE 8.8} & \cellcolor[HTML]{FCFBD8}\LARGE 7.1\textsubscript{1.1} & \cellcolor[HTML]{FCFBD8}\LARGE 4.1\textsubscript{0.6} & \LARGE 16.9\textsubscript{5.3} & \cellcolor[HTML]{FCFBD8}\LARGE 11.1\textsubscript{5.3} & \multicolumn{1}{c|}{\cellcolor[HTML]{FCFBD8}\LARGE 5.0\textsubscript{3.3}} & \multicolumn{1}{c|}{\LARGE 23.3} & \LARGE 28.6\textsubscript{9.9} & \LARGE 17.9\textsubscript{6.6} \\ \bottomrule
\end{tabular}
\endgroup
}
\caption{Main results}
\label{tab:main}
\end{table}

The proposed method achieves \textbf{SoTA overall accuracy improvements} on all tasks except for RTE across all ICL settings. As for RTE, due to its simplicity as a binary classification task, all three methods obtain similar accuracy and COBias improvements. Meanwhile, it achieves \textbf{SoTA COBias performance on general domain tasks}. In general, its COBias is on par with the current best methods, with average COBias over all seven tasks falling between that of DNIP and FuRud. On AGNews, DBpeida, and TREC, the proposed method achieves \textbf{both more overall accuracy improvements and more COBias reduction} than DNIP and FuRud across all ICL settings, demonstrating the superior performance gains achieved by ensembling class-level and sample-level debiasing. \textbf{Weak classes are elevated.} We show in Section \ref{subsec:scheme} that DCS powers up the low-accuracy classes. We observe that weaker classes are often corrected by rules, while stronger classes are most likely corrected by weights, further highlighting the need of an ensemble method.

\begin{table}[H]
\resizebox{\textwidth}{!}{%
\begin{tabular}{@{}lcccccc@{}}
\toprule
Dataset & \# Classes & \begin{tabular}[c]{@{}c@{}}\# Classes that \\ use weight \\ corrections\end{tabular} & \begin{tabular}[c]{@{}c@{}}\# Classes that \\ use membership \\ corrections\end{tabular} & \begin{tabular}[c]{@{}c@{}}Ratio of memberhsip\\ corrections to weight\\ corrections\end{tabular} & \begin{tabular}[c]{@{}c@{}}\#Classes that \\ use membership \\ corrections \\ in seed0/1/2 run\end{tabular} & \begin{tabular}[c]{@{}c@{}}Remarks on any weakest class\\ that use membership\\ corrections\end{tabular} \\ \midrule
\rowcolor[HTML]{DFF9FB} 
AGN & 4 & 3.7 & 0.3 & 0.08 & 0/1/0 & \begin{tabular}[c]{@{}c@{}}The weakest class \textit{Sports} from seed1 run,\\ Acc: 0.75 (ICL) $\rightarrow$ 0.98 (corrected)\end{tabular} \\
DBP & 14 & 12.3 & 1.7 & 0.14 & 2/2/1 & - \\
\rowcolor[HTML]{DFF9FB} 
SST & 5 & 4.3 & 0.7 & 0.16 & 1/0/1 & \begin{tabular}[c]{@{}c@{}}The weakest class \textit{Terrible} from seed0 run, \\ Acc: 0.21 (ICL) $\rightarrow$ 0.54 (corrected)\\ The weakest class \textit{Terrible} from seed2 run, \\ Acc: 0.09 (ICL) $\rightarrow$ 0.64 (corrected)\end{tabular} \\
TREC & 6 & 5 & 1 & 0.20 & 0/2/1 & \begin{tabular}[c]{@{}c@{}}The weakest class \textit{Entity} from seed2 run,\\ Acc: 0.04 (ICL) $\rightarrow$ 0.76 (corrected)\end{tabular} \\
\rowcolor[HTML]{DFF9FB} 
RTE & 2 & 0.7 & 1.3 & 1.86 & 1/2/1 & \begin{tabular}[c]{@{}c@{}}The weakest class \textit{False} from seed1 run,\\ Acc: 0.54 (ICL) $\rightarrow$ 0.78 (corrected)\end{tabular} \\
DDI & 5 & 2.7 & 2.3 & 0.85 & 3/2/2 & \begin{tabular}[c]{@{}c@{}}The weakest class \textit{Negative} from seed0 run, \\ Acc: 0 (ICL) $\rightarrow$ 0.73 (corrected)\\ The weakest class \textit{Negative} from seed1 run, \\ Acc: 0 (ICL) $\rightarrow$ 0.46 (corrected)\\ The weakest class \textit{Negative} from seed2 run, \\ Acc: 0.02 (ICL) $\rightarrow$ 0.93 (corrected)\end{tabular} \\
\rowcolor[HTML]{DFF9FB} 
PMQA & 3 & 2.7 & 0.3 & 0.11 & 0/1/0 & \begin{tabular}[c]{@{}c@{}}The weakest class \textit{Yes} from seed1 run,\\ Acc: 0.38 (ICL) $\rightarrow$ 0.62 (corrected)\end{tabular} \\ \bottomrule
\end{tabular}
}
\caption{Overall probability correction scheme ($N$-shot)}
\label{tab:scheme}
\end{table}

\noindent \textbf{Annealing Time}\\
\begin{wrapfigure}[12]{R}{0.4\textwidth}
  \begin{center}
\includegraphics[width=0.4\textwidth]{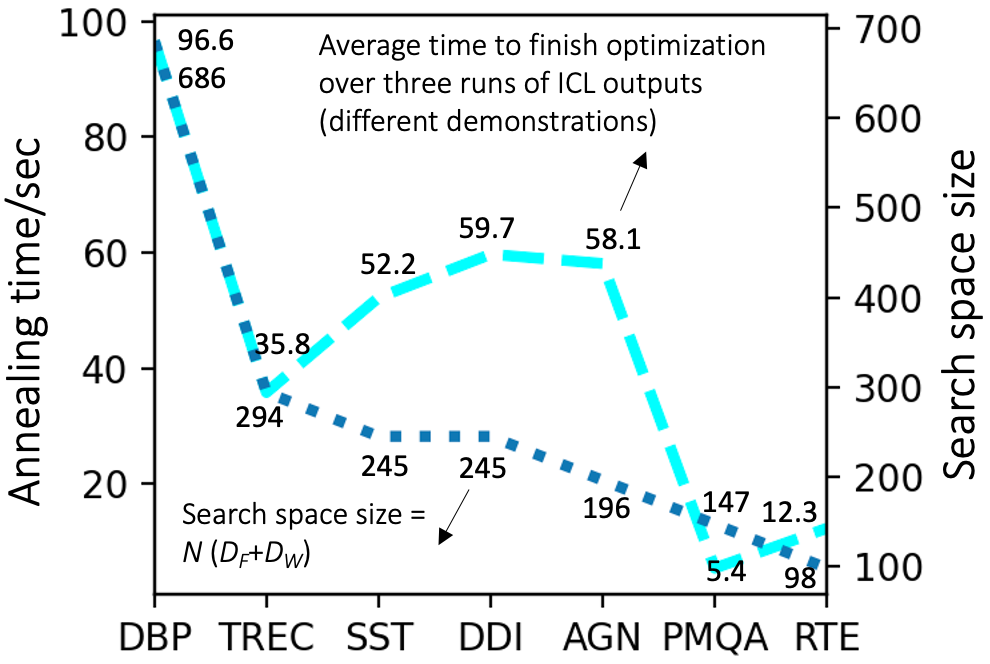}
  \end{center}
  \caption{Task annealing time}
  \label{fig:anntime}
\end{wrapfigure}
Figure \ref{fig:anntime} plots the time used to optimize correction functions (left axis) and the search space size (right axis) for each task. For fair comparisons, we fix the weight scale for weight correction to 30 alongside 19 triangular membership functions, making a total of 49 possible correction functions and $49\cdot N$ possible assignments across all variables. DCS is computationally efficient where all tasks finish optimization within 150 outer loops, ranging from a maximum of 96.6 seconds to a minimum of 5.4 seconds. Notably, SST-5, DDI, and AGNews have a relatively longer time due to having a larger optimization set (Appendix \ref{appdix:setup}). In general, although larger optimization sets and search spaces need more time to converge, the annealing steps needed for a task can be further tuned to achieve cost-effective gains on downstream applications.


\subsection{Overall Probability Correction Scheme}
\label{subsec:scheme}
The overall probability correction scheme demonstrates the ratio of class-level corrections to sample-level corrections among all classes. Which type of correction should be used for a class is automatically determined by our method. Recall that: \textit{Class-level correction}: Probabilities of the same output class are corrected uniformly using a same correction weight across all samples. It operates at the class level, not accounting for sample-based differences. \textit{Sample-level correction}: This is a more refined correction using membership correction functions. Instead of a single weight, a unique mapping function is applied for each class. This allows varying degrees of correction across different samples - some probabilities may be adjusted significantly, others minimally, and some may remain unchanged - capturing the nuances of individual samples, leading to more precise adjustments. By combining them, our method ensures a balanced and adaptive correction, enhancing performances.

We exemplify the overall probability correction scheme with the $N$-shot case in Table \ref{tab:scheme}. Numbers of classes are averaged over three runs of different demonstrations. As expected, most classes select class-level weight corrections. Membership corrections are less selected, but weakest classes benefit greatly from them, where 6 out of 7 tasks have selected membership corrections to correct their weakest classes (at least once over three runs). This demonstrates the necessity of enabling sample-level corrections to boost the performance of weak classes, and letting the optimization decide if sample-level corrections are needed.

\subsection{Larger Models on DDI (Biomedical-domain)}
\label{subsec:largermodel}

\begin{wraptable}{r}{6.8cm}
\centering
\resizebox{0.45\textwidth}{!}{%
\begin{tabular}{@{}lcccccccc@{}}
\toprule
\multicolumn{1}{l|}{} & \multicolumn{1}{c|}{} & \multicolumn{1}{c|}{} & \multicolumn{1}{c|}{} & \multicolumn{5}{c}{Class Acc.} \\ \cmidrule(l){5-9} 
\multicolumn{1}{l|}{\multirow{-2}{*}{Model}} & \multicolumn{1}{c|}{\multirow{-2}{*}{Acc.}} & \multicolumn{1}{c|}{\multirow{-2}{*}{CB}} & \multicolumn{1}{c|}{\multirow{-2}{*}{\begin{tabular}[c]{@{}c@{}}Prompt\\ Demon.\end{tabular}}} & Neg. & Eff. & Mech. & Adv. & Int. \\ \midrule
\multicolumn{9}{c}{\textit{Llama-2-13B}} \\ \midrule
\multicolumn{1}{l|}{} & \multicolumn{1}{c|}{} & \multicolumn{1}{c|}{} & \multicolumn{1}{c|}{seed0} & \cellcolor[HTML]{E4F7FC}0.0 & \cellcolor[HTML]{E4F7FC}0.0 & \cellcolor[HTML]{E4F7FC}100.0 & \cellcolor[HTML]{E4F7FC}0.0 & \cellcolor[HTML]{E4F7FC}2.3 \\
\multicolumn{1}{l|}{} & \multicolumn{1}{c|}{} & \multicolumn{1}{c|}{} & \multicolumn{1}{c|}{seed1} & \cellcolor[HTML]{E4F7FC}0.0 & \cellcolor[HTML]{E4F7FC}91.4 & \cellcolor[HTML]{E4F7FC}0.3 & \cellcolor[HTML]{E4F7FC}17.7 & \cellcolor[HTML]{E4F7FC}39.6 \\
\multicolumn{1}{l|}{\multirow{-3}{*}{ICL}} & \multicolumn{1}{c|}{\multirow{-3}{*}{7.2}} & \multicolumn{1}{c|}{\multirow{-3}{*}{45.6}} & \multicolumn{1}{c|}{seed2} & \cellcolor[HTML]{E4F7FC}0.0 & \cellcolor[HTML]{E4F7FC}0.8 & \cellcolor[HTML]{E4F7FC}84.1 & \cellcolor[HTML]{E4F7FC}54.8 & \cellcolor[HTML]{E4F7FC}88.5 \\
\multicolumn{1}{l|}{} & \multicolumn{1}{c|}{} & \multicolumn{1}{c|}{} & \multicolumn{1}{c|}{seed0} & \cellcolor[HTML]{FEE7E7}32.5 & \cellcolor[HTML]{FEE7E7}38.1 & \cellcolor[HTML]{FEE7E7}32.1 & \cellcolor[HTML]{FEE7E7}29.0 & \cellcolor[HTML]{FEE7E7}32.3 \\
\multicolumn{1}{l|}{} & \multicolumn{1}{c|}{} & \multicolumn{1}{c|}{} & \multicolumn{1}{c|}{seed1} & \cellcolor[HTML]{FEE7E7}62.1 & \cellcolor[HTML]{FEE7E7}30.6 & \cellcolor[HTML]{FEE7E7}32.8 & \cellcolor[HTML]{FEE7E7}42.1 & \cellcolor[HTML]{FEE7E7}31.3 \\
\multicolumn{1}{l|}{\multirow{-3}{*}{+DCS}} & \multicolumn{1}{c|}{\multirow{-3}{*}{\begin{tabular}[c]{@{}c@{}}52.9\\ \color{red}{\tiny (+45.7)}\color{black}\end{tabular}}} & \multicolumn{1}{c|}{\multirow{-3}{*}{\begin{tabular}[c]{@{}c@{}}16.4\\ \color{red}{\tiny (-29.2)}\color{black}\end{tabular}}} & \multicolumn{1}{c|}{seed2} & \cellcolor[HTML]{FEE7E7}80.7 & \cellcolor[HTML]{FEE7E7}12.8 & \cellcolor[HTML]{FEE7E7}8.6 & \cellcolor[HTML]{FEE7E7}6.3 & \cellcolor[HTML]{FEE7E7}11.5 \\ \midrule
\multicolumn{9}{c}{\textit{Llama-2-70B}} \\ \midrule
\multicolumn{1}{l|}{} & \multicolumn{1}{c|}{} & \multicolumn{1}{c|}{} & \multicolumn{1}{c|}{seed0} & \cellcolor[HTML]{E4F7FC}0.1 & \cellcolor[HTML]{E4F7FC}86.9 & \cellcolor[HTML]{E4F7FC}3.0 & \cellcolor[HTML]{E4F7FC}4.1 & \cellcolor[HTML]{E4F7FC}19.8 \\
\multicolumn{1}{l|}{} & \multicolumn{1}{c|}{} & \multicolumn{1}{c|}{} & \multicolumn{1}{c|}{seed1} & \cellcolor[HTML]{E4F7FC}8.7 & \cellcolor[HTML]{E4F7FC}64.2 & \cellcolor[HTML]{E4F7FC}3.0 & \cellcolor[HTML]{E4F7FC}18.1 & \cellcolor[HTML]{E4F7FC}56.3 \\
\multicolumn{1}{l|}{\multirow{-3}{*}{ICL}} & \multicolumn{1}{c|}{\multirow{-3}{*}{8.8}} & \multicolumn{1}{c|}{\multirow{-3}{*}{36.0}} & \multicolumn{1}{c|}{seed2} & \cellcolor[HTML]{E4F7FC}0.5 & \cellcolor[HTML]{E4F7FC}64.4 & \cellcolor[HTML]{E4F7FC}6.6 & \cellcolor[HTML]{E4F7FC}37.6 & \cellcolor[HTML]{E4F7FC}58.3 \\
\multicolumn{1}{l|}{} & \multicolumn{1}{c|}{} & \multicolumn{1}{c|}{} & \multicolumn{1}{c|}{seed0} & \cellcolor[HTML]{FEE7E7}69.9 & \cellcolor[HTML]{FEE7E7}34.4 & \cellcolor[HTML]{FEE7E7}19.9 & \cellcolor[HTML]{FEE7E7}18.6 & \cellcolor[HTML]{FEE7E7}47.9 \\
\multicolumn{1}{l|}{} & \multicolumn{1}{c|}{} & \multicolumn{1}{c|}{} & \multicolumn{1}{c|}{seed1} & \cellcolor[HTML]{FEE7E7}63.6 & \cellcolor[HTML]{FEE7E7}23.6 & \cellcolor[HTML]{FEE7E7}39.4 & \cellcolor[HTML]{FEE7E7}29.0 & \cellcolor[HTML]{FEE7E7}35.4 \\
\multicolumn{1}{l|}{\multirow{-3}{*}{+DCS}} & \multicolumn{1}{c|}{\multirow{-3}{*}{\begin{tabular}[c]{@{}c@{}}63.7\\ \color{red}{\tiny (+54.9)}\color{black}\end{tabular}}} & \multicolumn{1}{c|}{\multirow{-3}{*}{\begin{tabular}[c]{@{}c@{}}24.3\\ \color{red}{\tiny (-11.7)}\color{black}\end{tabular}}} & \multicolumn{1}{c|}{seed2} & \cellcolor[HTML]{FEE7E7}80.5 & \cellcolor[HTML]{FEE7E7}16.1 & \cellcolor[HTML]{FEE7E7}26.2 & \cellcolor[HTML]{FEE7E7}18.6 & \cellcolor[HTML]{FEE7E7}33.3 \\ \bottomrule
\end{tabular}
}
\caption{Significant DDI Performance gains on both Llama-2-13B and Llama-2-70B.}
\label{tab:70b}
\end{wraptable}

Similar to its smaller variant Llama-2-13B, 1-shot ICL on Llama-2-70B also manifests class accuracy imbalances. Especially on the biomedical domain DDI classification, using 1-shot prompting, Llama-2-13B's average ICL accuracy on DDI is 7.2\% with COBias 45.6\% as shown in Table \ref{tab:main}, while Llama-2-70B obtains 8.8\% accuracy with COBias 36.0\% as shown in Table \ref{tab:70b}, showing that the larger variant is still highly prone to biased predictions on biomedical tasks. This suggests that the class accuracy imbalance may not be effectively alleviated by increasing model sizes, and the proposed post-hoc corrections can exactly help with the situation. Due to effectively correcting probabilities for low-accuracy classes as shown by Table \ref{tab:70b}, our method demonstrates significant accuracy enhancements on DDI, across both small and large model variants. For \textbf{more Llama-2-70B performance gains}, detailed scores can be found in Appendix \ref{appdix:70b}.

\subsection{Ablations}
Some may contend that more overall accuracy enhancements can be compensated by less COBias reduction. While tradeoffs between overall accuracy and COBias generally exist, relaxing COBias reduction (resulting in larger COBias score) can not guarantee higher overall accuracy (Table \ref{tab:ablation}), showing the need of a dedicated $Z\textsuperscript{COBias}$ in the objective function.

\begin{table}[ht]
\centering
\resizebox{\textwidth}{!}{%
\begin{tabular}{@{}l|c|cccccccccccccc@{}}
\toprule
 &  & \multicolumn{2}{c}{AGN} & \multicolumn{2}{c}{DBP} & \multicolumn{2}{c}{SST} & \multicolumn{2}{c}{TREC} & \multicolumn{2}{c}{RTE} & \multicolumn{2}{c}{DDI} & \multicolumn{2}{c}{PMQA} \\ \cmidrule(l){3-16} 
\multirow{-2}{*}{Prompt} & \multirow{-2}{*}{Obj.} & Acc. & CB. & Acc. & CB. & Acc. & CB. & Acc. & CB. & Acc. & CB. & Acc. & CB. & Acc. & CB. \\ \midrule
 & Full & 88.2 & 6.2 & {\color[HTML]{00009B} 94.7\color{red}{\tiny ($\uparrow$6.1)}} & {\color[HTML]{00009B} 6.0\color{red}{\tiny ($\downarrow$10.2)}} & 50.2 & 23.7 & {\color[HTML]{00009B} 81.5\color{red}{\tiny ($\uparrow$12.9)}} & {\color[HTML]{00009B} 12.8\color{red}{\tiny ($\downarrow$23.1)}} & 74.4 & 5.0 & 52.9 & 16.4 & 63.3 & 35.2 \\
 & Z\textsuperscript{Err} & 88.4 & 7.8 & {\color[HTML]{34CDF9} 92.8\color{red}{\tiny ($\uparrow$4.2)}} & {\color[HTML]{34CDF9} 8.6\color{red}{\tiny ($\downarrow$7.6)}} & 51.4 & 50.9 & {\color[HTML]{34CDF9} 80.6\color{red}{\tiny ($\uparrow$12.0)}} & {\color[HTML]{34CDF9} 16.7\color{red}{\tiny ($\downarrow$19.2)}} & 75.7 & 12.8 & 82.2 & 39.8 & 71.3 & 58.8 \\
\multirow{-3}{*}{1-shot} & \multicolumn{1}{l|}{Z\textsuperscript{Err}+Z\textsuperscript{PMI}} & 88.5 & 7.5 & {\color[HTML]{3166FF} 94.1\color{red}{\tiny ($\uparrow$5.5)}} & {\color[HTML]{3166FF} 7.5\color{red}{\tiny ($\downarrow$8.7)}} & 51.7 & 53.0 & {\color[HTML]{3166FF} 81.1\color{red}{\tiny ($\uparrow$12.5)}} & \multicolumn{1}{l}{{\color[HTML]{3166FF} 15.1\color{red}{\tiny ($\downarrow$20.8)}}} & 75.8 & 13.0 & 82.9 & 40.0 & 71.1 & 59.1 \\
 & Full & 88.8 & 6.7 & {\color[HTML]{00009B} 96.3\color{red}{\tiny ($\uparrow$2.7)}} & {\color[HTML]{00009B} 4.7\color{red}{\tiny ($\downarrow$4.3)}} & 54.2 & 11.7 & {\color[HTML]{00009B} 79.7\color{red}{\tiny ($\uparrow$21.0)}} & {\color[HTML]{00009B} 14.0\color{red}{\tiny ($\downarrow$21.4)}} & 76.1 & 1.6 & 57.1 & 21.0 & 71.4 & 43.4 \\
 & Z\textsuperscript{Err} & 88.9 & 7.8 & {\color[HTML]{34CDF9} 95.0\color{red}{\tiny ($\uparrow$1.4)}} & {\color[HTML]{34CDF9} 7.1\color{red}{\tiny ($\downarrow$1.9)}} & 54.2 & 41.6 & {\color[HTML]{34CDF9} 79.5\color{red}{\tiny ($\uparrow$20.8)}} & {\color[HTML]{34CDF9} 14.6\color{red}{\tiny ($\downarrow$20.8)}} & 76.2 & 13.7 & 82.8 & 40.0 & 76.7 & 58.8 \\
\multirow{-3}{*}{5-shot} & \multicolumn{1}{l|}{Z\textsuperscript{Err}+Z\textsuperscript{PMI}} & 88.9 & 7.9 & {\color[HTML]{3166FF} 95.9\color{red}{\tiny ($\uparrow$2.3)}} & {\color[HTML]{3166FF} 5.5\color{red}{\tiny ($\downarrow$3.5)}} & 54.9 & 41.4 & {\color[HTML]{3166FF} 78.9\color{red}{\tiny ($\uparrow$20.2)}} & {\color[HTML]{3166FF} 21.0\color{red}{\tiny ($\downarrow$14.4)}} & 76.9 & 10.2 & 82.9 & 40.2 & 76.8 & 58.8 \\
 & Full & 88.9 & 7.1 & {\color[HTML]{00009B} 96.8\color{red}{\tiny ($\uparrow$1.7)}} & {\color[HTML]{00009B} 4.1\color{red}{\tiny ($\downarrow$2.8)}} & 54.2 & 16.9 & {\color[HTML]{00009B} 83.4\color{red}{\tiny ($\uparrow$16.4)}} & {\color[HTML]{00009B} 11.1\color{red}{\tiny ($\downarrow$27.1)}} & 76.7 & 5.0 & 62.4 & 28.6 & 60.3 & 17.9 \\
 & Z\textsuperscript{Err} & 88.8 & 7.5 & {\color[HTML]{34CDF9} 96.3\color{red}{\tiny ($\uparrow$1.2)}} & {\color[HTML]{34CDF9} 5.1\color{red}{\tiny ($\downarrow$1.8)}} & 54.4 & 37.4 & {\color[HTML]{34CDF9} 81.9\color{red}{\tiny ($\uparrow$14.9)}} & {\color[HTML]{34CDF9} 18.4\color{red}{\tiny ($\downarrow$19.8)}} & 76.9 & 17.0 & 82.8 & 40.1 & 76.0 & 60.5 \\
\multirow{-3}{*}{N-shot} & \multicolumn{1}{l|}{Z\textsuperscript{Err}+Z\textsuperscript{PMI}} & 89.0 & 7.5 & {\color[HTML]{3166FF} 96.5\color{red}{\tiny ($\uparrow$1.4)}} & {\color[HTML]{3166FF} 4.8\color{red}{\tiny ($\downarrow$2.1)}} & 54.8 & 40.8 & {\color[HTML]{3166FF} 83.3\color{red}{\tiny ($\uparrow$16.3)}} & {\color[HTML]{3166FF} 15.3\color{red}{\tiny ($\downarrow$22.9)}} & 77.5 & 15.7 & 82.9 & 40.1 & 76.1 & 60.7 \\ \bottomrule
\end{tabular}
}
\caption{Z\textsuperscript{Err} objective ablations; average scores over three random seeds are reported. Improvements over ICL scores are shown by ($\uparrow$) and ($\downarrow$).}
\label{tab:ablation}
\end{table}

To see this, we ablate the overall error term $Z\textsuperscript{Err}$ from the objective function and optimize solely for $Z\textsuperscript{Err}$. On DBpedia and TREC, DCS with only $Z\textsuperscript{Err}$ did not obtain a higher accuracy than the DCS with the full objective function. This is because the optimization without an explicit $Z\textsuperscript{COBias}$ term may favor higher-accuracy classes which are easier to improve, and becomes ``lazy'' to improve classes that have low accuracies initially. Applying the learned correction functions will result in a good overall test accuracy with fair COBias. On the other hand, explicitly setting $Z\textsuperscript{COBias}$ in the objective function can help enforce fair improvements for all classes while optimizing overall accuracy. It could bring more accuracy gains for lower-accuracy classes, and thus obtaining even higher overall test accuracy with lower COBias in some cases, especially when there are only a few low-accuracy classes among many higher-accuracy classes initially, like DBpedia and TREC. This suggests that balancing class accuracies can help with overall accuracy gains. Additionally, we ablate both accuracy objectives $Z\textsuperscript{Err}$ and $Z\textsuperscript{PMI}$, and optimize for these two terms without $Z\textsuperscript{COBias}$. On DBpedia and TREC, its performance generally falls between optimizing for $Z\textsuperscript{Err}$ and the full objective, further suggesting the importance of $Z\textsuperscript{COBias}$ in reaching optimal solutions.

\subsection{Discussions}
We visualize the \textbf{ablation of sample-level and class-level rebalancing} in Appendix \ref{appdix:vis}, demonstrating that sample-level rebalancing significantly elevates weak classes, while class-level rebalancing steadily balances the stronger ones. Moreover, \textbf{regarding more model families}, we provide experimental results with Gemma-2-2B in Appendix \ref{appdix:gemma}, which obtain consistent improvements similar to the Llama-2 cases, demonstrating that our method is applicable to LLMs of varied sizes and families. Lastly, with respect to \textbf{the time and cost of applying the DCS method in real-world scenarios}, we elaborate below. 

\textbf{Accessibility}: DCS is model-agnostic, requiring only output probabilities, making it particularly suitable for large or closed LLMs where only logits/output probabilities are accessible. 

\textbf{Cost of post-hoc optimization}: The post-hoc, offline optimization of DCS requires no model architecture modification and prompt engineering. The annealing time averages 46 seconds (computed from Figure \ref{fig:anntime}). Hyperparameter tuning can be done efficiently because most hyperparameters are pre-configured in practice, with only a few scalar values (e.g., $\beta, \tau$) requiring tuning. Moreover, since the class-level correction method DNIP demonstrates low-data optimization capability, we'd infer that DCS is amenable to low-data scenarios. These make DCS particularly valuable in specialized applications where fairness and accuracy are paramount.

\textbf{Negligible inference overhead}: The learned correction functions are reused on-the-fly with mere milliseconds in prediction, introducing virtually no computation overhead or latency. 

\textbf{Scalability}: Our method exhibits linear scaling with the number of classes ($N$). At each temperature in simulated annealing, the number of searches remains within several multiples of the search space size $N(D_F+D_W)$, ensuring practical feasibility across small and large classification tasks.

\section{Conclusion and Limitations}
\label{sec:concl}
This paper proposes DCS, a post-hoc ensemble debiasing framework to correct ICL probabilities across class and sample level. It boosts low-accuracy classes, while keeping the strong ones. It achieves SoTA accuracy on seven benchmark text classification tasks, while effectively mitigating class performance imbalances. The resulting probability correction scheme clearly shows that combining membership correction and weight correction is essential to elevate weak classes. For limitations and future work, the post-hoc correction framework can be extended to other useful scenarios, such as text generation. Combining DCS into the decoding process can help rectify the probabilities of top candidates for the first token (or first several tokens) to lead a better, less biased response. More future directions are discussed in Appendix \ref{appdix:morefuture}.

\section*{Acknowledgments}
Yang You's research group is being sponsored by NUS startup grant (Presidential Young Professorship), Singapore MOE Tier-1 grant, ByteDance grant, ARCTIC grant, SMI grant (WBS number: A-8001104-00-00), and Alibaba grant.

\bibliography{output}{}
\bibliographystyle{unsrtnat}

\newpage
\appendix
\section{Discussion on Solution Methods for NIP}
\label{appdix:bnb}
Typical solutions for combinatorial optimization problems include the branch-and-bound (B\&B) algorithm, which is useful for solving mixed-integer linear programs \citep{bnb}. Solvers are made into both open-source software such as CVXPY \citep{diamond2016cvxpy,agrawal2018rewriting} or SCIP \citep{scip}, and commercial solvers like Gurobi \citep{gurobi} or LINGO \citep{lingo}.

We need to point out that B\&B and other related algorithms are not very suitable to search solutions for the proposed nonlinear integer programming based optimization model, which contain discontinuous functions and is inherently non-differentiable. Relaxation to subproblems is needed if B\&B has to be used. Relaxation can be done by creating Lagrange multipliers and Lagrangian functions for each constraint, which is often easier to solve than the original nonlinear integer programming problem, because only continuous variables are involved \citep{luenberger2008,50yearsip}. The relaxed problem finds the optimal value of the Lagrangian function, which is a lower bound for the original problem, and it is used as a bound in the B\&B search. However, solving the subproblems efficiently poses challenges, as B\&B is generally not a polynomial-time algorithm.

Instead, we can opt for metaheuristics \citep{hussain2019metaheuristic}, such as simulated annealing \citep{kirkpatrick1983,Cerny}, to find the optimal solution. Although heuristic search can get slower with increasing search space size \citep{jia2019}, its computational complexity is within polynomial time \citep{cobias}. To enable application of our proposed method to tasks with even more classes, we adopt simulated annealing (SA) to solve our model. 

In SA, we define an initial temperature $T$ and an annealing schedule including the cooling rate and the stopping criteria. In each iteration, a neighborhood solution is generated by perturbing the current solution. The algorithm calculates the objective function value for the new solution (Step \textit{Evaluate Z} in Fig. \ref{fig:overall}). If the new solution is better, accept it; if the new solution is worse, accept it with a probability determined by the Metropolis criterion. Then the temperature is reduced according to the cooling rate, until convergence. In addition, the initial solution can also be randomly generated, where \(N\) numbers are randomly selected from \(\mathbb{D} = \{1, \dots, D_F+D_W\}\) as the initial values for the selection variables \(\xi_j\)'s. 

\section{List of notations}
\label{appdix:notations}
The parameters, sets and decision variables used throughout this paper are listed in Table \ref{tab:listofn}.

\begin{table}[htbp]
\centering
\begin{tabular}{@{}ll@{}}
\toprule
\textbf{Symbol} & \textbf{Description} \\ 
\midrule
Parameters \\
$p_m$ & The ICL output token's class probabilities of the $m$-th instance in a dataset \\
$\omega_{k}$ & The $k$-th weight coefficient where $k \in \{1,...,D_F\}$ \\
$\mu_{k}$ & The $k$-th membership function where $k \in \{D_F+1,...,D_F+D_W\}$ \\
\midrule
Sets \\
$\mathbb{R}^{\text{Weight}}$ & The set of $D_W$ weight coefficients, equally spaced discrete values from 0 to 1 \\
$\mathbb{R}^{\text{Membership}}$ & The set of $D_F$ triangular membership functions \\
$\mathbb{D}$ & The set of selection indices \\
\midrule
Variables \\
$y_m, \hat{y}_m$ & The label and prediction of the $m$-th instance \\
$\xi_j$ & The integer selection variable for class $j$ \\
\bottomrule
\end{tabular}
\caption{List of notations}
\label{tab:listofn}
\end{table}

\section{Details on the Objective Function}
\label{appdix:obj}
$Z^{\text{Err}}, Z^{\text{COBias}},Z^{\text{PMI}}$ respectively follows from the three terms of Eq. 5 in \citep{cobias}, which include:

\textit{The overall error rate for $M$ optimization instances:}
\begin{equation}
  Z^{\text{Err}} = \frac{1}{M} \displaystyle \sum_{m=1}^{M} \mathbbm{1}\{\hat{y}_m \ne y_m\}, \text{ where } \hat{y}_m = \argmax_{n \in \{1,\dots,N\}}  p'_{mn}, m=1,\dots,M  
\end{equation}

\textit{COBias of all class pairs:}
\begin{equation}
   Z^{\text{COBias}}={N\choose 2}^{-1} \sum_{i=1}^{N-1} \sum_{j=i+1}^{N} \bigg | A_{c_i} - A_{c_j} \bigg | \text{, where $A_{c*}$ is class $c*$'s accuracy}
\end{equation}

\textit{PMI between ground-truth instances of class $j$ and predictions of class $j$:}
\begin{equation}
    Z^{\text{PMI}}=-\sum_{j=1}^{N} \textrm{PMI}_j
\end{equation}
where
\begin{equation}
     \textrm{PMI}_j= \textrm{PMI}(\hat{S}^{j}, S^j) = \log \frac{f(\hat{S}^{j}, S^j)}{f(\hat{S}^{j})f(S^j)}, \text{ for } j=1,\dots,N
\end{equation}
Here, $\hat{S}^{j}$ and $S^j$ denote the instances of prediction $j$ and true label $j$ respectively, $f(\hat{S}^{j})$ is the ratio between the number of instances with prediction $j$ and the total number of instances, similarly, $f({S}^j)$ is the ratio between the number of instances labeled $j$ and the total number of instances, $f(\hat{S}^{j}, S^j)$ is the ratio between the number of correct predictions of class $j$ and the total number of instances.

\section{Experimental Setups}
\label{appdix:setup}
For each evaluation task, we tune the weight scale for weight correction, $\beta$ and $\tau$ used in the objective function, and the simulated annealing algorithm's number of outer loops, inner loops, the initial temperature, and stopping criteria (stopping temperature, the maximal number of solutions generated, and the maximal number of accepted solutions) on a development set. The optimization/development set split is 0.95/0.05. Dataset preprocessing follows DNIP \citep{cobias}, and their statistics are shown in Table \ref{tab:setup}.

\begin{table}[ht]
\resizebox{\textwidth}{!}{%
\begin{tabular}{@{}lccl@{}}
\toprule
Dataset & \begin{tabular}[c]{@{}c@{}}Optimization\\ set size \\ (incl. dev)\end{tabular} & \begin{tabular}[c]{@{}c@{}}Evaluation\\ set size\end{tabular} & \multicolumn{1}{c}{Prompt} \\ \midrule
\rowcolor[HTML]{ECF4FF} 
AGN & 10,000 & 5,000 & \begin{tabular}[c]{@{}l@{}}Please classify the news articles into the categories of \\ World, Sports, Business, and Technology.\textbackslash{}n\textbackslash{}n\\ Article: {[}Input{]}\\ Answer:\end{tabular} \\
DBP & 10,000 & 5,000 & \begin{tabular}[c]{@{}l@{}}Please classify the documents based on whether they are \\ about a Company, School, Artist, Athlete, Politician, \\ Transportation, Building, Nature, Village, Animal, \\ Plant, Album, Film, or Book.\textbackslash{}n\textbackslash{}n\\ Article: {[}Input{]}\\ Answer:\end{tabular} \\
\rowcolor[HTML]{ECF4FF} 
SST & 8,544 & 2,210 & \begin{tabular}[c]{@{}l@{}}Review: {[}Input{]}\\ Sentiment:\end{tabular} \\
TREC & 5,452 & 500 & \begin{tabular}[c]{@{}l@{}}Please classify the questions based on whether their answer\\ type is a Number, Location, Person, Description, Entity, \\ or Abbreviation.\textbackslash{}n\textbackslash{}n\\ Question: {[}Input{]}\\ Answer Type:\end{tabular} \\
\rowcolor[HTML]{ECF4FF} 
RTE & 2,490 & 277 & \begin{tabular}[c]{@{}l@{}}{[}Input premise{]}\\ question:\\ {[}Input hypothesis{]}\\ True or False?\\ answer:\end{tabular} \\
DDI & 10,000 & 5,716 & \begin{tabular}[c]{@{}l@{}}Please choose a most suitable answer from Negative, Effect, \\ Mechanism, Advice, or Interaction, for the drug-drug\\ interaction relation between the @drug\$ pair in the \\ following description.\textbackslash{}n\textbackslash{}n\\ Description: {[}Input{]}\\ Answer:\end{tabular} \\
\rowcolor[HTML]{ECF4FF} 
PMQA & 1,000 & 500 & \begin{tabular}[c]{@{}l@{}}Please choose a most suitable answer from yes, no, or maybe,\\ for the following question given a context.\textbackslash{}n\textbackslash{}n\\ Context: {[}Input context{]}\\ Question: {[}Input question{]} yes, no, or maybe?\\ Answer:\end{tabular} \\ \bottomrule

\end{tabular}
}
\caption{Dataset statistics}
\label{tab:setup}
\end{table}

\section{Full Performance Comparisons on Llama-2-70B}
\label{appdix:70b}
Our method effectively applies to larger LLM variants, showcased by Llama-2-70B. Debiasing results using DCS on 1-shot ICL outputs in Table \ref{tab:appdix_70b} show consistent improvements in average overall accuracy across three seeds of demonstrations, outperforming DNIP and FuRud. Our method achieves better COBias reduction than the other two methods on general domain tasks, and better accuracy with comparable COBias reduction on biomedical domain tasks. The takeaway is, increasing LLM sizes without post-hoc corrections may not always be an effective way to resolve output class accuracy imbalances, which further validates the necessity of the proposed post-hoc correction method.

\begin{table}[ht]
\centering
\resizebox{\textwidth}{!}{%
\begingroup
    \renewcommand{\arraystretch}{2}
\begin{tabular}{@{}lcccccccccccccccccccc@{}}
\toprule
\multicolumn{1}{l|}{} & \multicolumn{10}{c|}{\Huge Acc} & \multicolumn{10}{c}{\Huge COBias} \\ \cmidrule(l){2-21} 
\multicolumn{1}{l|}{} & \multicolumn{1}{c|}{} & \multicolumn{6}{c|}{\LARGE General Domain} & \multicolumn{3}{c|}{\LARGE Biomedical Domain} & \multicolumn{1}{c|}{} & \multicolumn{6}{c|}{\LARGE General Domain} & \multicolumn{3}{c}{\LARGE Biomedical Domain} \\ \cmidrule(lr){3-11} \cmidrule(l){13-21} 
\multicolumn{1}{l|}{\multirow{-3}{*}{\Huge Method}} & \multicolumn{1}{c|}{\multirow{-2}{*}{\LARGE All}} & \multicolumn{1}{c|}{\LARGE Avg} & \LARGE AGN &\LARGE  DBP & \LARGE SST & \LARGE TREC & \multicolumn{1}{c|}{\LARGE RTE} & \multicolumn{1}{c|}{\LARGE Avg} &\LARGE  DDI & \multicolumn{1}{c|}{\LARGE PMQA} & \multicolumn{1}{c|}{\multirow{-2}{*}{\LARGE All}} & \multicolumn{1}{c|}{\LARGE Avg} & \LARGE AGN & \LARGE DBP & \LARGE SST & \LARGE TREC & \multicolumn{1}{c|}{\LARGE RTE} & \multicolumn{1}{c|}{\LARGE Avg} & \LARGE DDI & \LARGE PMQA \\ \midrule
\multicolumn{21}{c}{\LARGE \textit{1-shot ICL + post-hoc debiasing}} \\ \midrule
\multicolumn{1}{l|}{\LARGE ICL} & \multicolumn{1}{c|}{\LARGE 66.4} & \multicolumn{1}{c|}{\LARGE 76.2} & \LARGE 87.4\textsubscript{2.2} & \LARGE 94.1\textsubscript{1.0} & \LARGE 41.8\textsubscript{8.6} & \LARGE 79.3\textsubscript{2.5} & \multicolumn{1}{c|}{\LARGE 78.6\textsubscript{2.6}} & \multicolumn{1}{c|}{\LARGE 41.7} & \LARGE 8.8\textsubscript{3.7} & \multicolumn{1}{c|}{\LARGE 74.5\textsubscript{2.3}} & \multicolumn{1}{c|}{\LARGE 31.8} & \multicolumn{1}{c|}{\LARGE 25.1} & \LARGE 14.5\textsubscript{5.6} & \LARGE 8.3\textsubscript{1.2} & \LARGE 54.7\textsubscript{1.4} & \LARGE 22.4\textsubscript{2.0} & \multicolumn{1}{c|}{\LARGE 25.5\textsubscript{7.6}} & \multicolumn{1}{c|}{\LARGE 48.8} & \LARGE 36.0\textsubscript{2.1} & \LARGE 61.5\textsubscript{3.3} \\
\multicolumn{1}{l|}{\LARGE  +DNIP} & \multicolumn{1}{c|}{\LARGE 73.6} & \multicolumn{1}{c|}{\LARGE 79.0} & \LARGE 89.8\textsubscript{0.2} & \LARGE 96.5\textsubscript{0.5} & \LARGE 49.4\textsubscript{2.0} & \LARGE 79.3\textsubscript{4.0} & \multicolumn{1}{c|}{\LARGE 80.0\textsubscript{1.1}} & \multicolumn{1}{c|}{\LARGE 60.0} & \LARGE 52.5\textsubscript{14.4} & \multicolumn{1}{c|}{\LARGE 67.5\textsubscript{2.4}} & \multicolumn{1}{c|}{\cellcolor[HTML]{F9EAEA}\LARGE 17.3} & \multicolumn{1}{c|}{\LARGE 15.9} & \cellcolor[HTML]{F9EAEA}\LARGE 6.5\textsubscript{0.4} & \LARGE 3.8\textsubscript{0.3} & \LARGE 47.1\textsubscript{14.0} & \LARGE 17.3\textsubscript{4.8} & \multicolumn{1}{c|}{\cellcolor[HTML]{F9EAEA}\LARGE 4.6\textsubscript{1.8}} & \multicolumn{1}{c|}{\cellcolor[HTML]{F9EAEA}\LARGE 21.1} & \cellcolor[HTML]{F9EAEA}\LARGE 16.4\textsubscript{9.0} & \cellcolor[HTML]{F9EAEA}\LARGE 25.7\textsubscript{2.0} \\
\multicolumn{1}{l|}{\LARGE  +FuRud} & \multicolumn{1}{c|}{\LARGE 75.9} & \multicolumn{1}{c|}{\LARGE 79.0} & \LARGE 89.4\textsubscript{0.6} & \LARGE 96.2\textsubscript{0.8} & \LARGE 50.0\textsubscript{2.2} & \LARGE 79.5\textsubscript{3.8} & \multicolumn{1}{c|}{\LARGE 79.9\textsubscript{9.1}} & \multicolumn{1}{c|}{\LARGE 68.1} & \LARGE 60.0\textsubscript{11.8} & \multicolumn{1}{c|}{\LARGE 76.1\textsubscript{2.2}} & \multicolumn{1}{c|}{\LARGE 21.4} & \multicolumn{1}{c|}{\LARGE 15.0} & \LARGE 7.1\textsubscript{1.0} & \LARGE 3.8\textsubscript{0.6} & \cellcolor[HTML]{F9EAEA}\LARGE 40.1\textsubscript{8.3} & \LARGE 19.0\textsubscript{5.2} & \multicolumn{1}{c|}{\LARGE 4.8\textsubscript{1.5}} & \multicolumn{1}{c|}{\LARGE 37.5} & \LARGE 25.2\textsubscript{7.5} & \LARGE 49.7\textsubscript{9.4} \\
\multicolumn{1}{l|}{\LARGE +DCS} & \multicolumn{1}{c|}{\cellcolor[HTML]{F9EAEA}\LARGE 76.8} & \multicolumn{1}{c|}{\cellcolor[HTML]{F9EAEA}\LARGE 79.6} & \cellcolor[HTML]{F9EAEA}\LARGE 89.9\textsubscript{0.2} & \cellcolor[HTML]{F9EAEA}\LARGE 96.7\textsubscript{0.5} & \cellcolor[HTML]{F9EAEA}\LARGE 50.1\textsubscript{1.7} & \cellcolor[HTML]{F9EAEA}\LARGE 81.0\textsubscript{3.6} & \multicolumn{1}{c|}{\cellcolor[HTML]{F9EAEA}\LARGE 80.1\textsubscript{1.1}} & \multicolumn{1}{c|}{\cellcolor[HTML]{F9EAEA}\LARGE 70.0} & \cellcolor[HTML]{F9EAEA}\LARGE 63.7\textsubscript{6.2} & \multicolumn{1}{c|}{\cellcolor[HTML]{F9EAEA}\LARGE 76.2\textsubscript{2.4}} & \multicolumn{1}{c|}{\LARGE 20.7} & \multicolumn{1}{c|}{\cellcolor[HTML]{F9EAEA}\LARGE 14.3} & \LARGE 6.6\textsubscript{0.1} & \cellcolor[HTML]{F9EAEA}\LARGE 3.6\textsubscript{0.2} & \LARGE 40.3\textsubscript{8.2} & \cellcolor[HTML]{F9EAEA}\LARGE 15.6\textsubscript{5.6} & \multicolumn{1}{c|}{\LARGE 5.3\textsubscript{1.6}} & \multicolumn{1}{c|}{\LARGE 36.8} & \LARGE 24.3\textsubscript{5.5} & \LARGE 49.2\textsubscript{9.1} \\ \bottomrule
\end{tabular}
\endgroup
}
\caption{Quantitative evaluations on Llama-2-70B}
\label{tab:appdix_70b}
\end{table}

\section{Ablation of Sample-level and Class-level Rebalancing}
\label{appdix:vis}
\begin{figure*}[ht] 
    \centering
    \begin{subfigure}[b]{0.48\textwidth} 
        \centering
        \includegraphics[width=\linewidth]{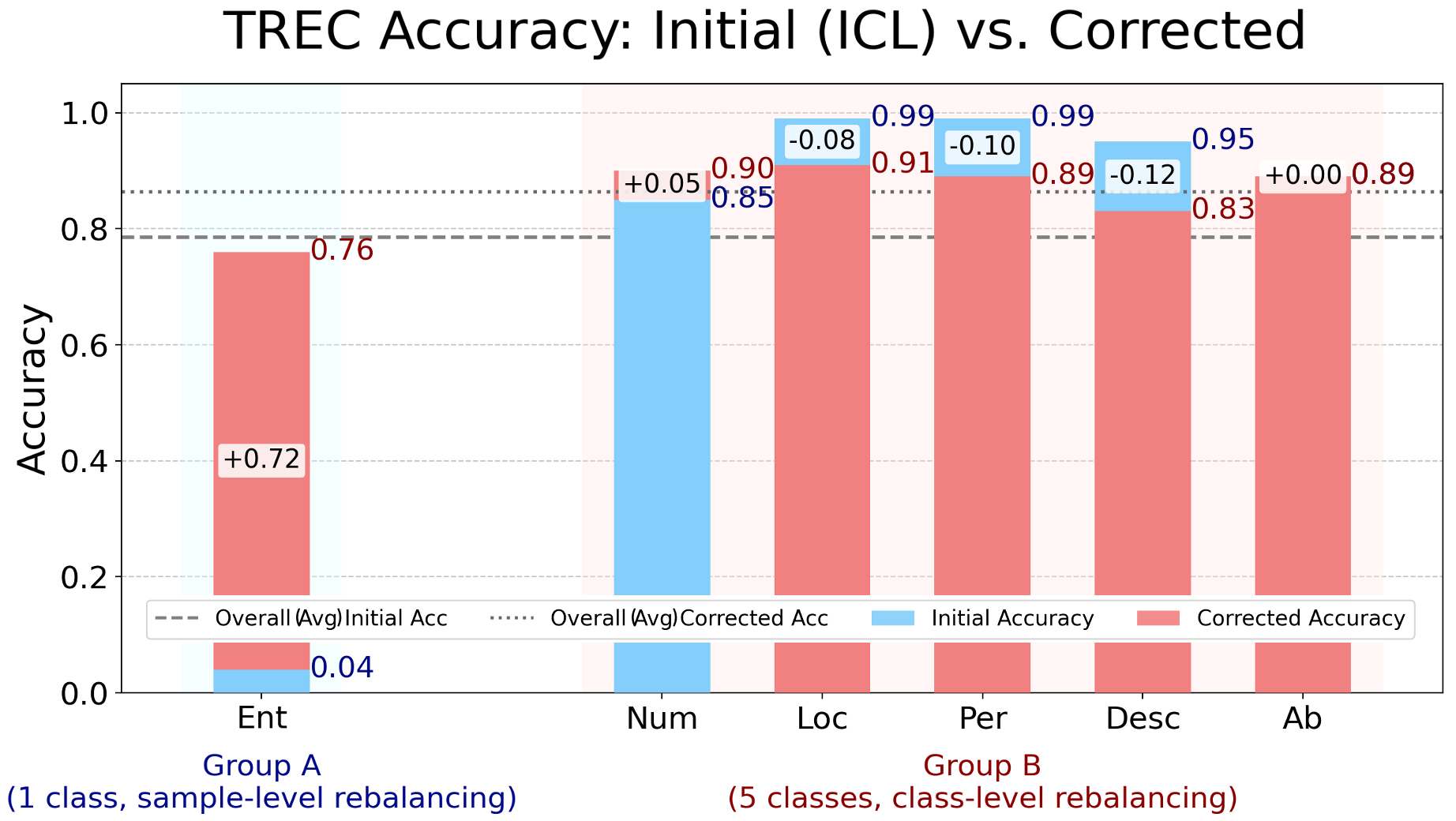}
        \caption{Accuracy by group of classes, TREC}
        \label{fig:trec}
    \end{subfigure}
    \hfill 
    \begin{subfigure}[b]{0.48\textwidth} 
        \centering
        \includegraphics[width=\linewidth]{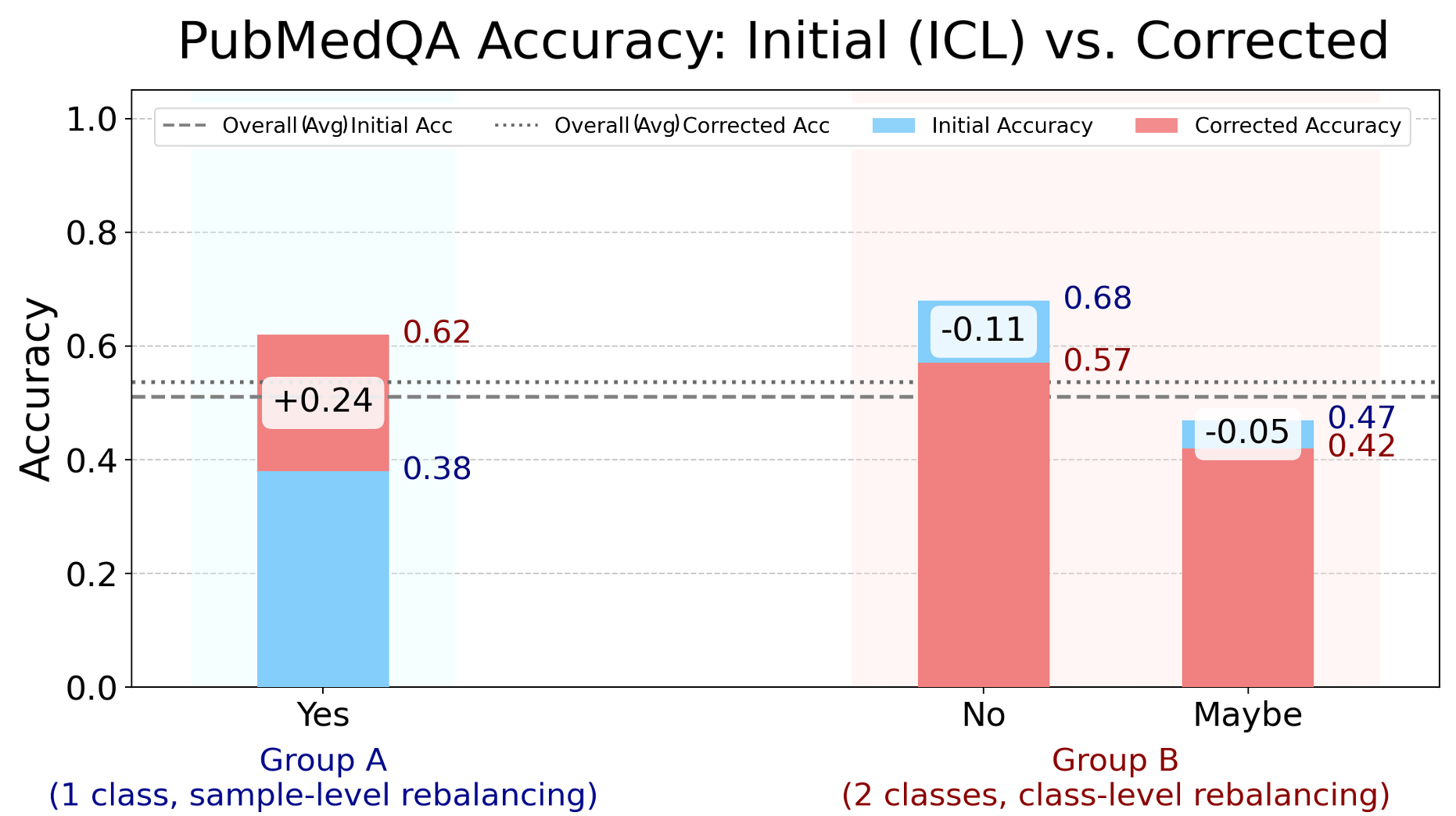}
        \caption{Accuracy by group of classes, PubMedQA}
        \label{fig:pub}
    \end{subfigure}
    \caption{Ablation of sample-/class-level rebalancing} 
    \label{fig:combined_figures} 
\end{figure*}

We visualize the ablation of sample-level and class-level rebalancing using TREC (general domain) and PubMedQA (biomedical domain) in Figure \ref{fig:trec} and \ref{fig:pub}. Both datasets show that sample-level rebalancing significantly elevates weak classes, while class-level rebalancing steadily balances the stronger ones. Ensembling yields overall gains in both metrics.

\section{Additional Performance Comparisons on Gemma-2-2B}
\label{appdix:gemma}

\begin{table}[ht]
\centering
\resizebox{\textwidth}{!}{%
\begingroup
    \renewcommand{\arraystretch}{2}
\begin{tabular}{@{}lcccccccccccccccccccc@{}}
\toprule
\multicolumn{1}{l|}{} & \multicolumn{10}{c|}{\Huge Acc} & \multicolumn{10}{c}{\Huge COBias} \\ \cmidrule(l){2-21} 
\multicolumn{1}{l|}{} & \multicolumn{1}{c|}{} & \multicolumn{6}{c|}{\LARGE General Domain} & \multicolumn{3}{c|}{\LARGE Biomedical Domain} & \multicolumn{1}{c|}{} & \multicolumn{6}{c|}{\LARGE General Domain} & \multicolumn{3}{c}{\LARGE Biomedical Domain} \\ \cmidrule(lr){3-11} \cmidrule(l){13-21} 
\multicolumn{1}{l|}{\multirow{-3}{*}{\Huge Method}} & \multicolumn{1}{c|}{\multirow{-2}{*}{\LARGE All}} & \multicolumn{1}{c|}{\LARGE Avg} & \LARGE AGN &\LARGE  DBP & \LARGE SST & \LARGE TREC & \multicolumn{1}{c|}{\LARGE RTE} & \multicolumn{1}{c|}{\LARGE Avg} &\LARGE  DDI & \multicolumn{1}{c|}{\LARGE PMQA} & \multicolumn{1}{c|}{\multirow{-2}{*}{\LARGE All}} & \multicolumn{1}{c|}{\LARGE Avg} & \LARGE AGN & \LARGE DBP & \LARGE SST & \LARGE TREC & \multicolumn{1}{c|}{\LARGE RTE} & \multicolumn{1}{c|}{\LARGE Avg} & \LARGE DDI & \LARGE PMQA \\ \midrule
\multicolumn{21}{c}{\LARGE \textit{1-shot ICL + post-hoc debiasing}} \\ \midrule

\multicolumn{1}{l|}{\LARGE  ICL} & \multicolumn{1}{c|}{\LARGE 53.4} & \multicolumn{1}{c|}{\LARGE 59.0} & \LARGE 79.9\textsubscript{3.4} & \LARGE 72.6\textsubscript{2.7} & \LARGE 27.6\textsubscript{4.3} & \LARGE 44.6\textsubscript{4.7} & \multicolumn{1}{c|}{\LARGE 70.3\textsubscript{6.6}} & \multicolumn{1}{c|}{\LARGE 39.4} & \LARGE 15.4\textsubscript{10.6} & \multicolumn{1}{c|}{\LARGE 63.4\textsubscript{1.0}} & \multicolumn{1}{c|}{\LARGE 38.5} & \multicolumn{1}{c|}{\LARGE 35.9} & \LARGE 25.7\textsubscript{6.9} & \LARGE 36.6\textsubscript{1.8} & \LARGE 42.0\textsubscript{1.5} & \LARGE 50.2\textsubscript{0.7} & \multicolumn{1}{c|}{\LARGE 25.2\textsubscript{13.1}} & \multicolumn{1}{c|}{\LARGE 44.8} & \LARGE 35.1\textsubscript{13.0} & \LARGE 54.4\textsubscript{3.2} \\
\multicolumn{1}{l|}{\LARGE  +DNIP} & \multicolumn{1}{c|}{\LARGE 65.1} & \multicolumn{1}{c|}{\LARGE 70.1} & \LARGE 86.9\textsubscript{1.1} & \LARGE 88.4\textsubscript{0.9} & \LARGE 33.2\textsubscript{7.9} & \LARGE 69.6\textsubscript{0.8} & \multicolumn{1}{c|}{\cellcolor[HTML]{F9EAEA}\LARGE 72.6\textsubscript{2.3}} & \multicolumn{1}{c|}{\LARGE 52.5} & \LARGE 41.5\textsubscript{18.2} & \multicolumn{1}{c|}{\LARGE 63.4\textsubscript{11.3}} & \multicolumn{1}{c|}{\cellcolor[HTML]{F9EAEA}\LARGE 19.9} & \multicolumn{1}{c|}{\cellcolor[HTML]{F9EAEA}\LARGE 15.9} & \LARGE \cellcolor[HTML]{F9EAEA}7.1\textsubscript{0.5} & \LARGE 14.0\textsubscript{1.0} & \LARGE \cellcolor[HTML]{F9EAEA}34.3\textsubscript{9.9} & \LARGE 21.4\textsubscript{2.8} & \multicolumn{1}{c|}{\cellcolor[HTML]{F9EAEA}\LARGE 2.6\textsubscript{1.8}} & \multicolumn{1}{c|}{\cellcolor[HTML]{F9EAEA}\LARGE 30.0} & \LARGE 18.3\textsubscript{7.2} & \LARGE \cellcolor[HTML]{F9EAEA}41.8\textsubscript{27.3} \\
\multicolumn{1}{l|}{\LARGE  +FuRud} & \multicolumn{1}{c|}{\LARGE 66.9} & \multicolumn{1}{c|}{\LARGE 71.6} & \LARGE 85.3\textsubscript{1.8} & \LARGE 88.4\textsubscript{1.7} & \LARGE 44.1\textsubscript{8.5} & \LARGE 68.5\textsubscript{0.6} & \multicolumn{1}{c|}{\LARGE 71.7\textsubscript{4.7}} & \multicolumn{1}{c|}{\LARGE 55.3} & \LARGE 43.0\textsubscript{18.9} & \multicolumn{1}{c|}{\LARGE 67.5\textsubscript{1.2}} & \multicolumn{1}{c|}{\LARGE 23.1} & \multicolumn{1}{c|}{\LARGE 18.3} & \LARGE 12.3\textsubscript{2.3} & \LARGE 12.7\textsubscript{3.6} & \LARGE 40.1\textsubscript{17.0} & \LARGE 21.6\textsubscript{1.2} & \multicolumn{1}{c|}{\LARGE 4.7\textsubscript{3.0}} & \multicolumn{1}{c|}{\LARGE 35.1} & \LARGE \cellcolor[HTML]{F9EAEA}18.2\textsubscript{10.9} & \LARGE 51.9\textsubscript{0.6} \\
\multicolumn{1}{l|}{\LARGE  +DCS (Ours)} & \multicolumn{1}{c|}{\cellcolor[HTML]{F9EAEA}\LARGE 69.2} & \multicolumn{1}{c|}{\cellcolor[HTML]{F9EAEA}\LARGE 74.5} & \LARGE \cellcolor[HTML]{F9EAEA}87.0\textsubscript{1.1} & \LARGE \cellcolor[HTML]{F9EAEA}92.5\textsubscript{0.9} & \LARGE \cellcolor[HTML]{F9EAEA}49.4\textsubscript{1.7} & \LARGE \cellcolor[HTML]{F9EAEA}70.9\textsubscript{2.5} & \multicolumn{1}{c|}{\cellcolor[HTML]{F9EAEA}\LARGE 72.6\textsubscript{2.3}} & \multicolumn{1}{c|}{\cellcolor[HTML]{F9EAEA}\LARGE 56.0} & \LARGE \cellcolor[HTML]{F9EAEA}44.1\textsubscript{0.3} & \multicolumn{1}{c|}{\cellcolor[HTML]{F9EAEA}\LARGE 67.9\textsubscript{2.3}} & \multicolumn{1}{c|}{\LARGE 23.3} & \multicolumn{1}{c|}{\LARGE 17.8} & \LARGE 7.7\textsubscript{1.0} & \LARGE \cellcolor[HTML]{F9EAEA}6.8\textsubscript{0.3} & \LARGE 50.7\textsubscript{0.9} & \LARGE \cellcolor[HTML]{F9EAEA}21.3\textsubscript{2.8} & \multicolumn{1}{c|}{\cellcolor[HTML]{F9EAEA}\LARGE 2.6\textsubscript{1.8}} & \multicolumn{1}{c|}{\LARGE 37.0} & \LARGE 22.4\textsubscript{4.2} & \LARGE 51.5\textsubscript{4.8} \\ \bottomrule
\end{tabular}
\endgroup
}
\caption{Quantitative evaluations on Gemma-2-2B}
\label{tab:appdix_gemma2b}
\end{table}
We additionally experimented with Gemma-2-2B, and obtained consistent improvements (Table \ref{tab:appdix_gemma2b}) similar to the Llama cases, showing that our method is applicable to LLMs of varied sizes and families.

\section{More Future Directions}
\label{appdix:morefuture}
This work deals with prediction biases when directly prompting LLMs. When LLMs or smaller LMs like BERT \citep{bert} are finetuned on tasks, which may result in further imbalanced class accuracies, post-hoc corrections can make class accuracies fairer. It would be valuable to understand how these output class probabilities differ from those by direct prompting, and enable custom extensions to the current framework to provide more targeted corrections. It would also be interesting to see how it may help with data-filtered or data-augmented finetuning. In terms of interpretability, those classes that apply class-level correction to any instance is interpretable only at the broader level, so further investigations on how to enforce more interpretability to weight corrections could be conducted.

\end{document}